\documentclass[final]{cvpr}

\usepackage{times}
\usepackage{epsfig}
\usepackage{graphicx}
\usepackage{amsmath}
\usepackage{amssymb}
\usepackage{multirow}
\usepackage{appendix}

\usepackage[pagebackref=true,breaklinks=true,colorlinks,bookmarks=false]{hyperref}

\def\cvprPaperID{6457} 
\def\confYear{CVPR 2021}

\begin{document}

\title{Detecting Human-Object Interaction via Fabricated Compositional Learning}


\author{Zhi Hou$^1$, Baosheng Yu$^1$, Yu Qiao$^{2,3}$, Xiaojiang Peng$^{4}$, Dacheng Tao$^1$ \\
$^1$ School of Computer Science, Faculty of Engineering, The University of Sydney, Australia \\
$^2$ 
Shenzhen Institute of Advanced Technology, Chinese Academy of Sciences \\
$^3$ Shanghai AI Laboratory \\
$^4$ Shenzhen Technology University \\
{\tt\small zhou9878@uni.sydney.edu.au, baosheng.yu@sydney.edu.au, yu.qiao@siat.ac.cn,} \\ {\tt\small pengxiaojiang@sztu.edu.cn, dacheng.tao@sydney.edu.au}


}

\maketitle

\begin{abstract}


Human-Object Interaction (HOI) detection, inferring the relationships between human and objects from images/videos, is a fundamental task for high-level scene understanding. However, HOI detection usually suffers from the open long-tailed nature of interactions with objects, while human has extremely powerful compositional perception ability to cognize rare or unseen HOI samples. Inspired by this, we devise a novel HOI compositional learning framework, termed as Fabricated Compositional Learning (FCL), to address the problem of open long-tailed HOI detection. Specifically, we introduce an object fabricator to generate effective object representations, and then combine verbs and fabricated objects to compose new HOI samples. With the proposed object fabricator, we are able to generate large-scale HOI samples for rare and unseen categories to alleviate the open long-tailed issues in HOI detection. Extensive experiments on the most popular HOI detection dataset, HICO-DET, demonstrate the effectiveness of the proposed method for imbalanced HOI detection and significantly improve the state-of-the-art performance on rare and unseen HOI categories. Code is available at \url{https://github.com/zhihou7/HOI-CL}.



\end{abstract}

\section{Introduction}

\begin{figure}[t]
\begin{center}
\includegraphics[width=0.48\textwidth]{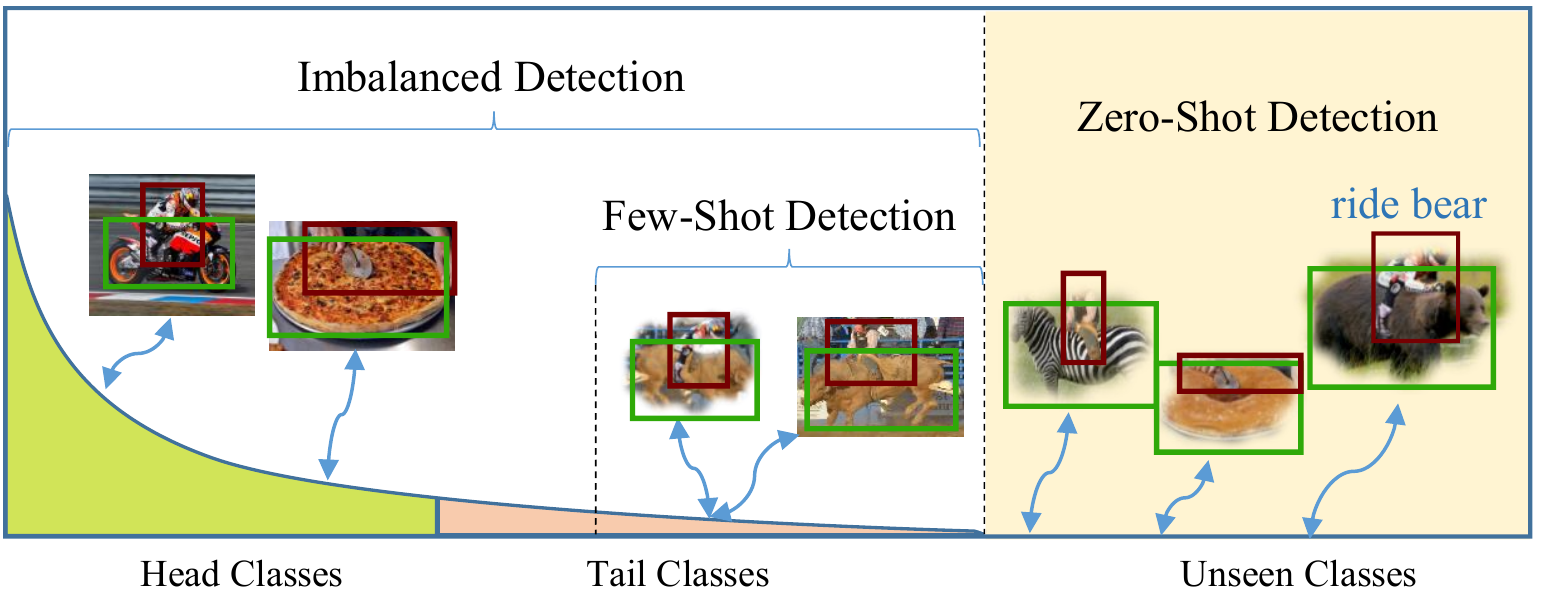}
\end{center}
   \caption{Open long-tailed HOI detection addresses the problem of imbalanced learning and zero-shot learning in a unified way. We propose to compose new HOIs for open long-tailed HOI detection. Specifically, the blurred HOIs, \eg, ``ride bear", are composite. See more examples in supplementary materials.}
\label{fig:open_long_tailed}
\end{figure}

Human-Object Interaction (HOI) detection, which aims to localize and infer relationships between human and objects in images/videos, $\left \langle human, verb, object \right \rangle$, is an essential step towards deeper scene and action understanding~\cite{chao2018learning, gao2018ican}. In real-world scenarios, long-tailed distributions are common for the data perceived by human vision system, \eg, actions/verbs and objects~\cite{liu2019large}. The combinatorial nature of HOI further highlights the issues of long-tailed distributions in HOI detection, while human can efficiently learn to recognize seen and even unseen HOIs from limited samples. An intuitive example of open long-tailed HOI detection is shown in Figure~\ref{fig:open_long_tailed}, in which one can easily recognize the unseen action ``ride bear'', nevertheless it never even happened. However, existing HOI detection approaches usually focus on either the head \cite{gao2018ican,liao2019ppdm,wang2018low}, the tail~\cite{xu2019learning} or unseen categories~\cite{shen2018scaling,preye2019detecting}, leaving the
problem of open long-tailed HOI detection poorly investigated.


Open long-tailed HOI detection falls into the category of the long-tailed zero-shot learning problem, which is usually referred into several isolated problems, including long-tailed learning \cite{japkowicz2002class, he2009learning}, few-shot learning \cite{fei2006one, vinyals2016matching}, zero-shot learning \cite{lampert2009learning}. To address the problem of imbalanced training data,  existing methods mainly focus on three strategies: 1) re-sampling \cite{drummond2003c4, han2005borderline, kang2019decoupling}; 2) re-weighted loss functions \cite{cui2019class, cao2019learning, hayat2019gaussian}; and 3) knowledge transfer \cite{wang2017learning, liu2019large, fei2006one, lampert2009learning, schonfeld2019generalized, frome2013devise}. Specifically, re-sampling and re-weighted loss functions are usually designed for imbalance problem, while knowledge transfer is introduced to relieve all the long-tailed \cite{wang2017learning}, few-shot \cite{snell2017prototypical}, and zero-shot problem~\cite{xian2018feature, frome2013devise}. Recently, two popular knowledge transfer methods have received increasing attention from the community, data generation~\cite{wang2017learning, wang2018low, xian2018feature, liu2019large, fei2006one, lampert2009learning, schonfeld2019generalized, keshari2020generalized} (transferring head/base classes to tail/unseen classes) and visual-semantic embedding~\cite{frome2013devise} (transferring from language knowledge). Along the first way, we address the problem of open long-tailed HOI detection from the perspective of HOI generation.

\begin{figure}[t]
\begin{center}
\includegraphics[width=.45\textwidth]{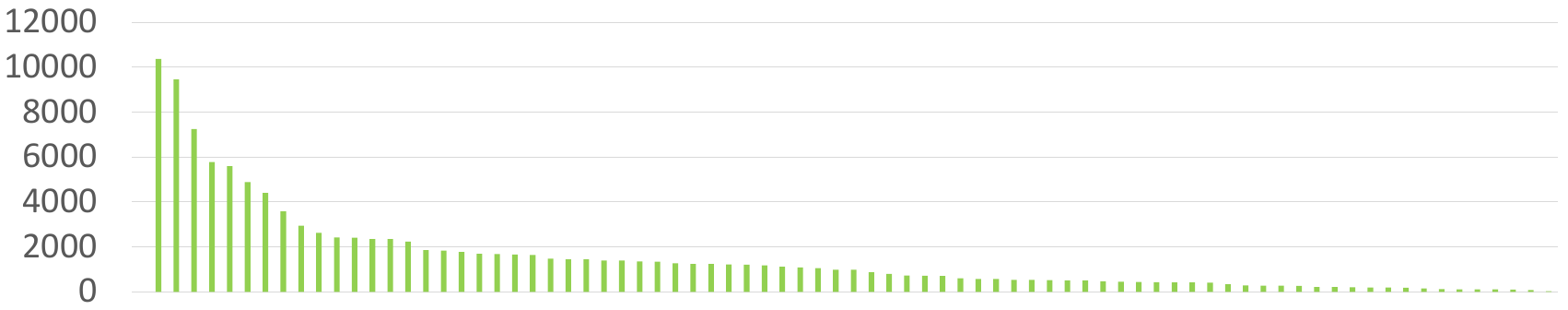}
\end{center}
  \caption{Illustration of distribution of the number of object box in HICO-DET dataset. The categories are sorted by the number of instances.}
\label{fig:obj_distribution}
\end{figure}

Unlike the samples in typical long-tailed zero-shot learning for visual recognition, each HOI sample is composed of a verb and an object, and different HOIs may share the same verb or object (\eg, ``ride bike'' and ``ride horse''). In cognitive science, human perceives concepts  as the compositions of shareable components~\cite{biederman1987recognition, hoffman1983parts} (\eg, verb and object in HOI), which indicates that human can conceive a new concept through a composition of existing components. Inspired by this, several zero-and few-shot HOI detection approaches have been proposed to enforce the factored primitive (verb and object) representation of the same primitive class to be similar among different HOIs, such as factorized model \cite{shen2018scaling, bansal2020detecting} and factor visual-language model \cite{xu2019learning, preye2019detecting, bansal2020detecting}. However, regularizing factor representation, \ie enforcing the same verb/object representation to be similar among different HOIs, is only sub-optimal for HOI detection. Recently, Hou \etal \cite{hou2020visual} present to compose novel HOI samples via combining decomposed verbs and objects between pair-wise images and within image. Nevertheless, it still remains a great challenge to compose massive HOI samples in each minibatch from images due to limited number of HOIs in each image, especially when the distribution of objects/verbs is also long-tailed. We demonstrate the distribution of the number of objects in Figure~\ref{fig:obj_distribution}.


The long-tailed distribution of objects/verbs makes it difficult to compose new HOIs from each mini-batch, significantly degrading the performance of compositional learning-based methods for rare and zero-shot HOI detection~\cite{hou2020visual}. Inspired by recent success of visual object representation generation~\cite{xian2018feature, hariharan2017low, wang2018low}, we thus apply fabricated object representation, instead of fabricated verb representation, to compose more balanced HOIs. We referred to the proposed compositional learning framework with fabricated object representation as Fabricated Compositional Learning or FCL. Specifically, we first extract verb representations from input images, and then design a simple yet efficient object fabricator to generate object representation. Next, the generated visual object features are further combined with the verb features to compose new HOI samples. With the proposed object fabricator, we are able to generate balanced objects for each verb within the mini-batch of training data as well as compose massive balanced HOI training samples.

The main contributions of this paper can be summarized as follows: 1) proposing to compose HOI samples for Open Long-Tailed HOI detection; 2) designing an object fabricator to generate objects for HOI composition; 3) significantly outperforming recent state-of-the-art methods on HICO-DET dataset among rare and unseen categories.


\begin{figure*}
\centering
\includegraphics[width=0.90\textwidth]{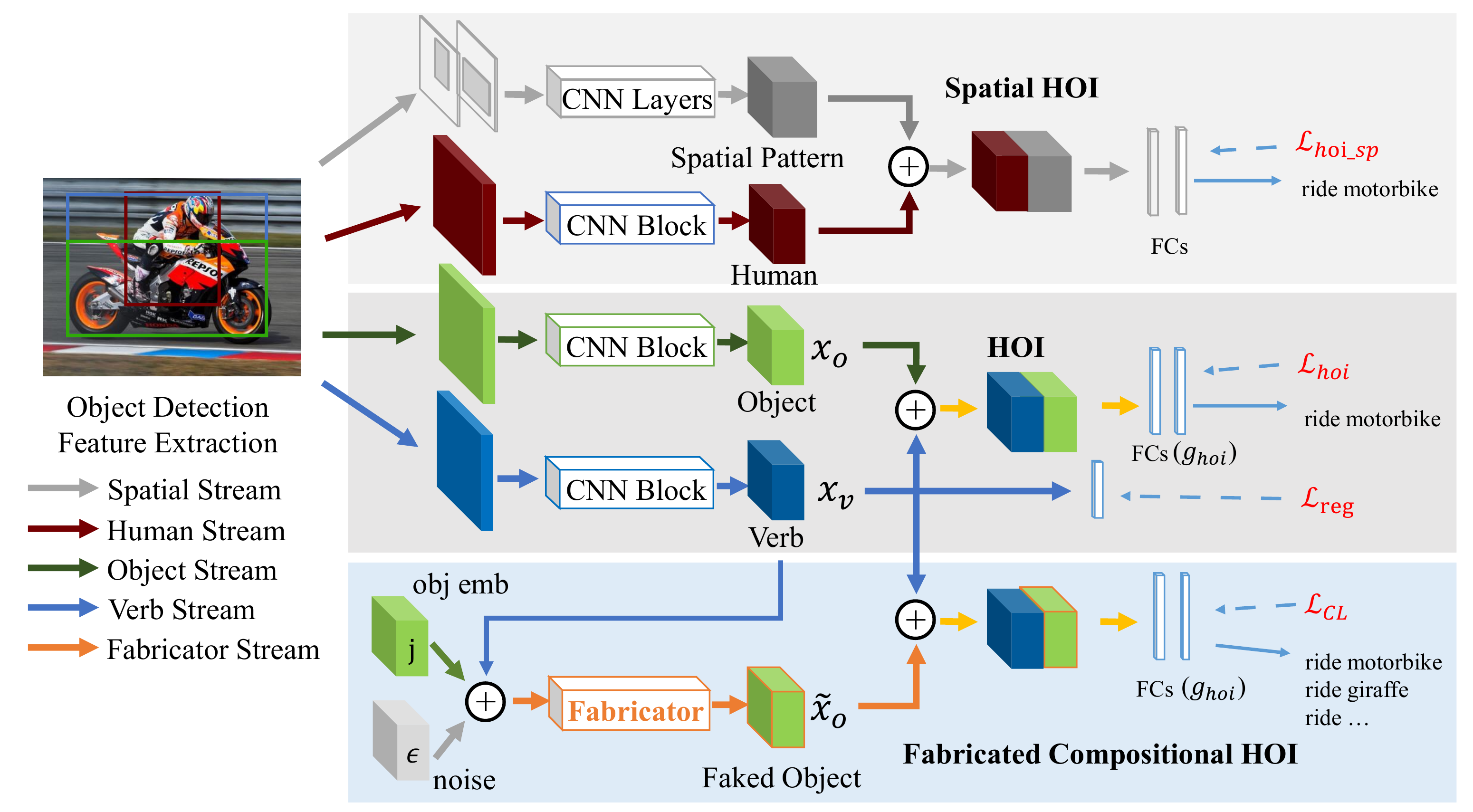}
\caption{An overview of the proposed multi-branch fabricated compositional learning framework for HOI detection. We first detect human and object with Faster-RCNN \cite{ren2015faster} from the image. Next, with ROI-Pooling and residual CNN blocks, we extract human features, verb features and object features. Meanwhile, an object identity embedding, verb feature and noise are input into Fabricator to generate fake object feature. Then, these features are fed into the following branches: individual spatial HOI branch, HOI branch and fabricated compositional HOI branch. Finally, HOI representations from HOI branch and fabricated branch are optimized by a shared FC-Classifier, while HOI representations from spatial branch are classified by an individual FC-Classifier. In fabricated compositional HOI branch, verb features are combined with fabricated objects to construct fabricated HOIs.}
\label{fig:pipeline}
\end{figure*}

\section{Related Works}

{\bf HOI Detection}.
HOI detection is essential for deeper scene and action understanding~\cite{chao2018learning}. Recent HOI detection approaches usually focus on representation learning \cite{gao2018ican, zhou2019relation, ulutan2020vsgnet, wang2019deep, wan2019pose}, zero/few-shot generalization \cite{shen2018scaling, xu2019learning, preye2019detecting, bansal2020detecting, hou2020visual}, and One-Stage HOI detection \cite{liao2019ppdm, wang2020learning}. Specifically, existing methods improve HOI representation learning by exploring the relationships among different features \cite{qi2018learning, zhou2019relation, ulutan2020vsgnet}, including pose information \cite{li2018transferable, wan2019pose, li2020detailed}, context \cite{gao2018ican, wang2019deep}, and human parts \cite{zhou2019relation}; Generalization methods for HOI detection mainly include visual-language model \cite{preye2019detecting, xu2019learning}, factorized model \cite{shen2018scaling, gupta2019no, ulutan2020vsgnet, bansal2020detecting}, and HOI composition \cite{hou2020visual}. Recently, Liao \etal \cite{liao2019ppdm} and Wang \etal \cite{wang2020learning} propose to detect the interaction point for HOI by heatmap-based localization \cite{newell2016stacked}. Wang \etal \cite{wang2020discovering} try to detect HOI with novel objects by leveraging human visual clues to localize interacting objects. However, existing HOI approaches usually fail to investigate the imbalance issue and zero-shot detection. Inspired by the factorized model \cite{shen2018scaling}, we propose to compose visual verb and fabricated objects to address the open long-tailed issue in HOI detection. Furthermore, according to whether detect the objects with a separated detector or not, existing HOI detection approaches can be divided into two categories: 1) one-stage \cite{shen2018scaling, liao2019ppdm, wang2020learning, gkioxari2018detecting} and two-stage \cite{gao2018ican, li2018transferable, zhou2019relation, ulutan2020vsgnet, wang2019deep, xu2019learning, bansal2020detecting, ulutan2020vsgnet}. Two-stage methods usually achieve better performance and our method falls into this category.

{\bf Compositional Learning}.
Irving Biederman illustrates that human representations of concepts are decomposable~\cite{biederman1987recognition}. Meanwhile, Lake \etal \cite{lake2017building} argue compositionality is one of the key blocks in a human-like learning system. Tokmakov \etal \cite{tokmakov2019learning} apply the compositional deep representation into few-shot learning. External knowledge graph and graph
convolutional networks in \cite{kato2018compositional} are used to compose verb-object pairs for HOI recognition. Recently, Hou \etal \cite{hou2020visual} propose a novel visual compositional learning framework to compose HOIs from image-pairs for HOI detection, failing to address the open and long-tailed issues. Therefore, we further compose verb and fake object representations for HOI detection.

{\bf Generalized Zero/Few-Shot Learning}.
Different from typical zero/few-shot learning \cite{fei2006one, lampert2009learning, vinyals2016matching}, generalized zero/few-shot learning \cite{xian2018zero} is a more realistic variant, since the performance is evaluated on both seen and unseen classes~\cite{schonfeld2019generalized, chao2016empirical}. The distribution of HOIs is naturally long-tailed \cite{chao2018learning}, \ie, most classes have a few training examples. Moreover, the open long-tailed HOI detection aims to handle the long-tailed, low-shot and zero-shot issue in a unified way. The long-tailed data distribution \cite{japkowicz2002class, he2009learning, huang2016learning} is one of challenging problem in visual recognition. Currently, re-sampling \cite{gupta2019lvis, kang2019decoupling}, specific loss \cite{lin2017focal, cui2019class, cao2019learning, hayat2019gaussian}, knowledge transfer \cite{wang2017learning, liu2019large}, and data generation~\cite{wang2018low,kumar2018generalized, xian2018feature, alfassy2019laso} are major strategies for imbalanced learning \cite{japkowicz2002class, he2009learning, huang2016learning}. To make full use of the composition characteristic of HOI, we aim to compose HOI samples by visual feature generation to relieve the open long-tailed issue in HOI detection. Recent feature generation methods \cite{kumar2018generalized, xian2018feature} mainly depend on Variational Autoencoder \cite{kingma2013auto} and Generative Adversarial Network \cite{goodfellow2014generative}, which usually suffer from the problem of model collapse \cite{salimans2016improved}. Wang \etal \cite{wang2018low} present a new method for low-shot learning that directly learns to hallucinate examples that are useful for classification. Similar to \cite{wang2018low}, we compose HOI samples with an object fabricator in an end-to-end optimization without using the adversarial loss.

\section{Method}

In this section, we first describe the multi-branch compositional learning framework for HOI detection. We then introduce the proposed fabricated compositional learning for open long-tailed HOI detection.


\subsection{Multi-branch HOI Detection}

HOI detection aims to find the interactions between human and different objects in a given image/video. Existing HOI detection methods~\cite{gao2018ican, li2018transferable, bansal2020detecting} usually contain two separated stages: 1) human and object detection; and 2) interaction detection. Specifically, we first use a common object detector, \eg, Faster R-CNN~\cite{ren2015faster}, to localize the positions and extract the features for both human and objects. According to the union of human and object bounding boxes, we then extract the verb feature from the feature map of backbone networks via the ROI-Pooling operation. Similar to~\cite{gao2018ican, gupta2019no, li2018transferable}, an additional stream for spatial pattern, \ie, spatial stream, is defined as the concatenation of human and object masks, \ie, the value in the human/object bounding box region is 1 and 0 elsewhere. As a result, we obtain several input streams from the first stage, \ie, human stream, object stream, verb stream, and spatial stream.

The input streams from the first stage then are used to construct different branches in the second stage: 1) \textbf{the spatial HOI branch}, which concatenates the spatial and the human streams to construct spatial HOI feature for HOI recognition; 2) \textbf{the HOI branch}, which concatenates the verb and the object streams; and 3) \textbf{the fabricated compositional branch}, which is based on a new stream, the fabricator stream, to generate fake object features for composing new HOIs. Specifically, the fabricated compositional branch generates novel HOIs by combining visual verb features and generated object features. The main multi-branch HOI detection framework is shown in Figure~\ref{fig:pipeline}, and we leave the details of the fabricated compositional branch in next section.

\subsection{Fabricated Compositional Learning}
\label{sec:fab}

The motivation of compositional learning is to decompose a model/concept into several sub-models/concepts, in which each sub-model/concept focuses on a specific task, and then all responses are coordinated and aggregated to make the final prediction~\cite{biederman1987recognition}. Recent compositional learning method for HOI detection considers each HOI as the combination of a verb and an object to compose new HOIs from objects and verbs within the mini-batch of training samples~\cite{kato2018compositional, hou2020visual}. However, existing compositional learning methods fail to address the problem of long-tailed distribution on objects.


To address the open long-tailed issue, we propose to generate balanced objects for each decoupled visual verb as follows. Formally, we denote $\mathbf{l}_{v}$ as the label of a verb $x_v$, $\mathbf{l}_{o}$ as the label of an object $x_o$ and $\mathbf{y}$ as the HOI label of $\left \langle x_v, x_o \right \rangle$.
Given another verb representation $\hat{x}_v$ (sharing the same label $\mathbf{l}_{v}$ with $x_v$), and another object representation $\hat{x}_o$ (sharing the same label $\mathbf{l}_{o}$ with $x_o$), regardless of the sources of the verb and object representations, an effective composition of verb and object should be
\begin{equation}
\label{eq:compose}
g_{hoi}(\hat{x}_{v}, \hat{x}_{o}) \approx g_{hoi}(x_v, x_o),
\end{equation}
where $g_{hoi}$ indicates the HOI classification network. By doing this, we can compose new verb-object pair $\left \langle \hat{x}_v, \hat{x}_o \right \rangle$, which  have similar semantic type $\mathbf{y}$ to the real pair $\left \langle x_v, x_o \right \rangle$, to relieve the scarcity of rare and unseen HOI categories. To generate effective verb-object pair $\left \langle \hat{x}_v, \hat{x}_o \right \rangle$, we regularize the verb representation $\hat{x}_v$ and object representation $\hat{x}_o$ such that same verbs/objects have similar feature representation.

Similar to previous approaches, such as factor visual-language joint embedding \cite{xu2019learning, preye2019detecting} and factorized model \cite{shen2018scaling, gupta2019no},  when $\hat{x}_v$ is similar to $x_v$ and $\hat{x}_o$ is similar to $x_o$, we then have that Equation~\eqref{eq:compose} can be generalized to HOI detection via the compositional branch. We refer to the proposed compositional learning framework with fabricated object representation as Fabricated Compositional Learning or FCL. We train the proposed method with composited HOI samples $\left \langle \hat{x}_v, \hat{x}_o \right \rangle$ in an end-to-end manner, and the overall loss function are defined as follows:


\begin{equation}
\label{eq:cl}
\mathcal{L} = \lambda_1 \mathcal{L}_{hoi} +  \lambda_2 \mathcal{L}_{CL} +  \lambda_3 \mathcal{L}_{reg} + \mathcal{L}_{hoi\_sp},
\end{equation}
where $\mathcal{L}_{reg}$ aims to regularize verb and object features, $\mathcal{L}_{CL}$ indicates a typical compositional learning loss function for the classification network $g_{hoi}$ with composite HOI samples $\left \langle \hat{x}_v, \hat{x}_o \right \rangle$ as the input, $\mathcal{L}_{hoi\_sp}$ is the loss for Spatial HOI Branch. $\lambda_1, \lambda_2, \lambda_3$ are the hyper-parameters to balance different loss functions. Specifically, object feature extracted from a pre-trained object detector backbone network (\ie Faster-RCNN \cite{ren2015faster}) are usually discriminative. Thus, we only regularize verb representation.

\begin{figure}
\centering
\includegraphics[width=0.48\textwidth]{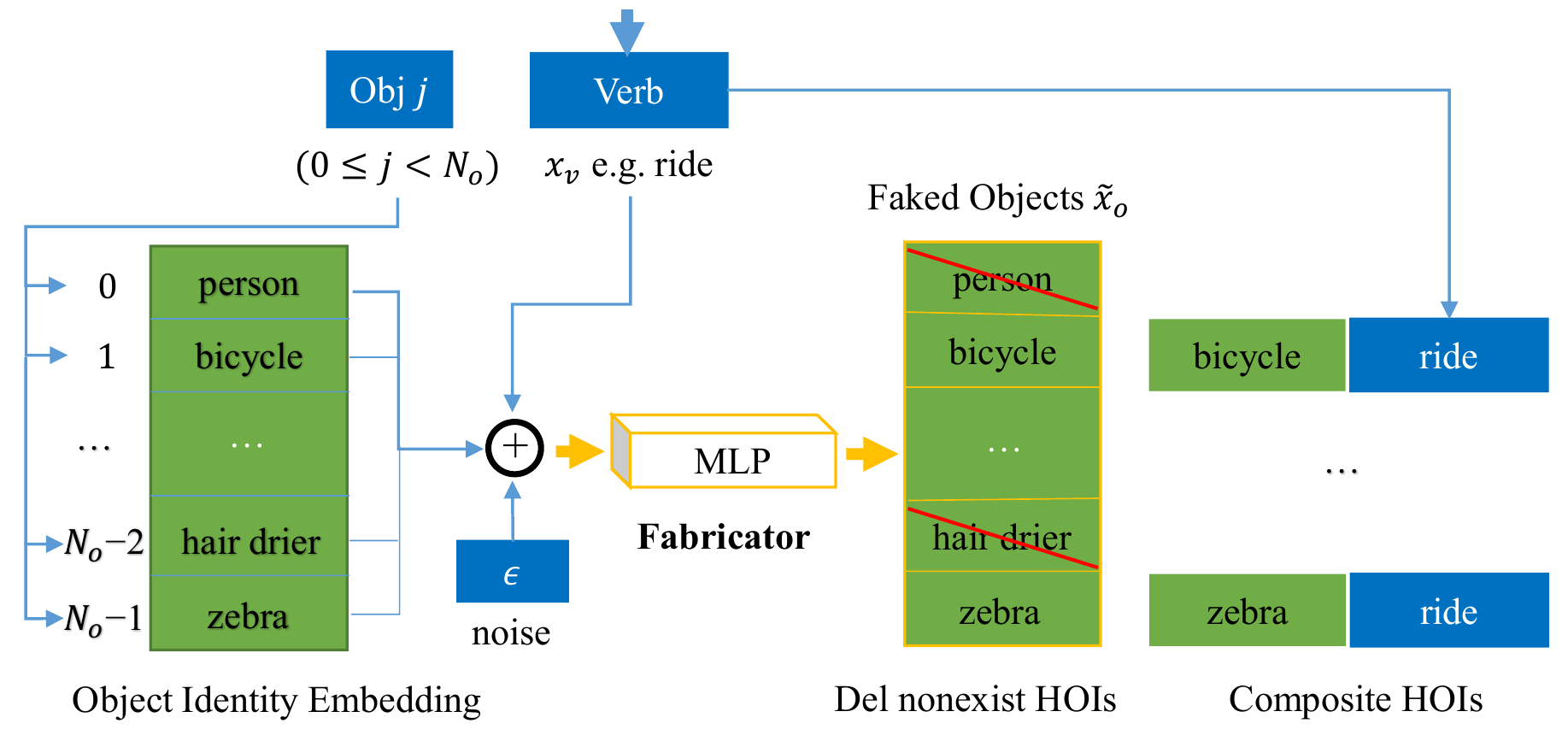}
\caption{For a given visual verb feature and each $j_{th}$ ($0\leq j <  N_o$), we firstly select the $j_{th}$ object identity embedding. Then, we concatenate verb feature, object embedding and Gaussian noise to input to fabricator for generating a fake object feature. We can fabricate $N_o$ objects for a verb feature. We finally remove nonexisting HOIs as described in Section 3.2.2.}
\label{fig:fabricating}
\end{figure}

\subsubsection{Object Generation}
The HOI is composed of a verb and an object, in which the verb is usually a very abstract notation compared to the object, making it difficult to directly generate verb features. Recent visual feature generation methods have demonstrated the effectiveness of feature generation for visual object recognition~\cite{wang2018low, xian2018feature}. Therefore, we devise an object fabricator to generate object feature representations for composing novel HOI samples.

The overall framework of object generation is shown in Figure~\ref{fig:fabricating}. Specifically, we maintain a pool of object identity embeddings, \ie, $v_{id}$. We provide three kinds of embeddings in supplementary material. In each HOI, the pose of the object is usually influenced by the human who is interacting the object \cite{zhang2020phosa}, and the person who is interacting with the object is firmly related to verb feature representation. Thus, for each extracted verb and the $j_{th}$ object ($0\leq j <  N_o$ and $N_o$ is the number of all different objects), we concatenate the $j_{th}$ object identity embedding $v_{id}^j$, the verb feature $x_v$ and a noise vector $\epsilon \sim \mathcal{N}(0,1)$, as the input of the object fabricator, \ie,
\begin{equation}
\hat{x}_o = f_{obj}(\{v_{id}^j, x_v, \epsilon\}),
\end{equation}
where $\hat{x}_o$ is the fake object feature and $f$ indicates the object fabricator network. Here, the noise $\epsilon$ is used to increase the diversity of generated objects. We then combine the fake object feature $\hat{x}_o$ and the verb $x_v$ to compose a new HOI sample $\left \langle x_v, \hat{x}_o \right \rangle$. Specifically, during training, both real HOIs and composite HOIs share the same HOI classification network $g_{hoi}$.

\subsubsection{Efficient HOI Composition}
To compose new HOIs from verb and object representations, we need to remove some infeasible composite HOIs (\eg, ``ride vase") as illustrated in Figure~\ref{fig:fabricating}. To avoid frequently checking the pair $(x_{v}, x_{o})$, we use an efficient HOI composition similar to~\cite{hou2020visual}. Specifically, the HOI label space is decoupled into verb and object spaces, \ie, the co-occurrence matrices $\mathbf{A}_v\in R^{N_v\times C}$ and $\mathbf{A}_o\in R^{N_o\times C}$, where $N_v$, $N_o$, and $C$ indicate the number of verbs, objects and HOI categories, respectively. Given an one-hot HOI label vector $\mathbf{y} \in R^{C}$, we then have the verb label vectors,
\begin{equation}
\label{eq:decompose}
\mathbf{l}_v = \mathbf{y} \mathbf{A}_v ^\mathsf{T} ,
\end{equation}
where $\mathbf{l}_v \in R^{N_v}$ can be a multi-hot vector with multiple verbs, \eg, $\left \langle \{hold, read\}, book \right \rangle$). Similarly, combining the verb $\mathbf{l}_v$ with all $N_o$ objects, we have the matrix $\mathbf{\hat{l}}_o \in R^{N_o \times N_o}$ as labels of all $N_o$ fake objects. Let $\mathbf{\hat{l}}_v \in R^{N_o \times N_v}$ denote the verb labels corresponding to fake object features $\mathbf{\hat{l}}_o$, the new interaction label can then be evaluated as follows,
\begin{equation}
\label{eq:label}
\hat{\mathbf{y}} = (\mathbf{\hat{l}}_o \mathbf{A}_o)~\&~(\mathbf{\hat{l}}_v\mathbf{A}_v),
\end{equation}
where $\&$ indicates the logical operation ``$\mathbf{and}$''.
Finally, the logical operation automatically filters out the infeasible HOIs since the labels of those infeasible HOIs are all-zero vectors in the label space.

\subsection{Optimization}
\label{sec:verb_loss}
\textbf{Training}. The verb feature contains the pose information of the object, making it difficult to jointly train the network with an object fabricator from scratch. Therefore, we introduce a step-wise training strategy for the long-tailed HOI detection. Firstly, we pre-train the network by $\mathcal{L}_{hoi}$, $\mathcal{L}_{hoi\_sp}$ and $\mathcal{L}_{reg}$ without the fabricator branch. Then, we fix the pre-trained model and train the randomly initialized object fabricator via the loss function for the fabricator branch $\mathcal{L}_{CL}$. Lastly, we jointly fine-tune all branches by $\mathcal{L}$ in an end-to-end manner. To avoid the bias to seen data in the first step, we optimize the network in one step for zero-shot HOI detection (See analysis in Section~\ref{sec:ab}).


\textbf{Inference}. The fabricated branch is only used in the training stage, \ie, we remove it during the inference stage. Similar to previous multi-branch methods~ \cite{gao2018ican, li2018transferable, hou2020visual}, for each human-object bounding box pair ($b_h$, $b_o$), the final HOI prediction  $S^c_{h,o}$ for each category $c \in 1, ..., C$, can be evaluated as follows,
\begin{equation}
\label{eq:final_score}
S^c_{h,o} = s_h\cdot s_o \cdot S^c_{sp} \cdot S^c_{hoi},
\end{equation}
where $s_h$ and $s_o$ indicate the object detection scores for the human and object, respectively.  $S^c_{sp}$ and $S^c_{hoi}$ are the scores from the Spatial branch and the HOI branch, respectively.

\section{Experiments}
In this section, we first introduce datasets and metrics, and then provide the details of the implementation of our method. Next, we present our experimental results compared with state-of-the-art approaches. Finally, we conduct ablation studies to validate the components in our work.

\subsection{Datasets and Metrics}

We adopt the largest HOI datasets HICO-DET \cite{chao2018learning}, which contains 47,776 images including 38,118 images for training and 9,658 images for testing. All 600 HOI categories are constructed from 80 object categories and 117 verb categories. HICO-DET provides more than 150k annotated human-object pairs. In addition, V-COCO is another small HOI dataset with 29 categories~\cite{gupta2015visual}. Considering that V-COCO mainly focuses to verb recognition and do not contain a severe long-tailed issue, we mainly evaluate the proposed method on HICO-DET. We also
illustrate the result on visual relation detection \cite{lu2016visual, zhan2019exploring}, which requires to detect the triplet (subject, predicate, object) in supplementary materials. We follow the evaluation settings in \cite{chao2018learning}, \ie a HOI prediction is a true positive if 1) both the human and object bounding boxes have IoUs larger than 0.5 with the reference ground truth bounding boxes; and 2) the HOI prediction is accurate.

\subsection{Implementation Details}


Similar to \cite{bansal2020detecting, hou2020visual}, our HOI detection model contains two separated stages: 1) we finetune the Faster R-CNN detector pre-trained on COCO~\cite{lin2014microsoft} using HICO-DET to detect the human and objects~\footnote{We use the Faster R-CNN detector implemented in detectron2 \cite{wu2019detectron2}.}; 2) we use the proposed FCL model for HOI classification. Specifically, all branches are two-layer MLP sigmoid classifiers with 2048-d input and 1024-d hidden units. Fabricator is a two-layer MLP. The $\mathcal{L}_{reg}$ is a sigmoid classifier for verb representation. $\mathcal{L}_{CL}$, $\mathcal{L}_{hoi}$ and $\mathcal{L}_{hoi\_sp}$ are binary cross entropy losses. $\mathbf{A}_v$ and $\mathbf{A}_o$ are set according to HOI dataset, and we can also set them by prior knowledge to detect more types of unseen HOIs. Besides, to prevent the fabricated HOIs from dominating the model optimization process, we randomly sample fabricated HOIs in each mini-batch to keep that the number of fabricated HOIs is not more than three times the number of non-fabricated HOIs. We train our network for one million iterations by SGD optimizer on the HICO-DET dataset with an initial learning rate of 0.01, a weight decay of 0.0005, and a momentum of 0.9. We set $\lambda_1$ as 2.0, $\lambda_2$ as 0.5 and $\lambda_3$ as 0.3, while we set 1 for the coefficient of $\mathcal{L}_{hoi\_sp}$. The hyper-parameters are ablated in supplementary materials. We jointly fine-tune the model with the object fabricator for ~500k iterations, and decay the initial learning rate 0.01 with a cosine annealing schedule. All our experiments on HICO-DET are conducted using TensorFlow~\cite{abadi2016tensorflow} on a single Nvidia GeForce RTX 2080Ti GPU. We evaluate V-COCO based on PMFNet \cite{wan2019pose} with two GPUs. We do not use auxiliary verb loss since there are only two kinds of objects on V-COCO. We set $\lambda_1$ as 1 and $\lambda_2$ as 0.25 on V-COCO.

\subsection{Comparison to Recent State-of-the-Arts}

Our method aims to relieve open long-tailed HOI detection. However current approaches usually focus on full categories, rare categories and unseen categories separately. In order to compare with state-of-the-art methods, we evaluate our method on long-tailed detection and generalized zero-shot detection separately. The HOI detection result is evaluated with mean average precision (mAP) (\%).

\subsubsection{Effectiveness for Zero-Shot HOI Detection}

There are different settings \cite{bansal2020detecting} for zero-shot HOI detection: 1) unseen composition; and 2) unseen object. Specifically, for the unseen composition setting, it indicates that the training data contains all factors (\ie, verbs and objects) but misses the verb-object pairs; for the unseen object setting, it requires to detect unseen HOIs, in which the object do not appear in the training data. For unseen composition HOI detection, similar to \cite{hou2020visual}, we select two groups of 120 unseen HOIs from tail preferentially (rare first) and from head preferentially (non-rare first) separately, which roughly compares the lowest and highest performances. As a result, we report our result in the following settings: Unseen (120 HOIs), Seen (480 HOIs), Full (600 HOIs) in the ``Default" mode on HICO-DET dataset. For a better comparison, we implement the factorized model \cite{shen2018scaling} under our framework for unseen composition zero-shot HOI detection. For unseen object HOI detection, we use the same HOI categories for unseen data as \cite{bansal2020detecting} (\ie randomly selecting 12 objects from the 80 objects and picking all HOIs containing there objects as unseen HOIs). Then, we report our results in the setting: Unseen (100 HOIs), Seen (500 HOIs), Full (600 HOIs).
 To compare with the contemporary work~\cite{hou2020visual}, we use the same object detection result released by~\cite{hou2020visual}. Here, our baseline method is the model without object fabricator, \ie, the compositional branch.

\setlength{\tabcolsep}{4pt}
\begin{table}[tp]
\caption{Comparison of zero-shot detection results of our proposed method. UC indicates unseen composition zero-shot HOI detection.
UO indicates unseen object zero-shot HOI detection. For better illustration, we choose the mean UC result of \cite{bansal2020detecting}.
  }
\label{table:zero_shot1}
\centering
\small
\begin{tabular}{@{}lcccccc@{}}
\hline
Method & Type & Unseen & Seen & Full \cr
\hline\hline
Shen \etal \cite{shen2018scaling} & UC & 5.62 & - & 6.26 \\
FG \cite{bansal2020detecting}  & UC & 11.31 & 12.74 & 12.45 \\
\hline
VCL \cite{hou2020visual} (rare first) & UC & 10.06 & 24.28 & 21.43 \\
Baseline (rare first)  & UC & 8.94 & 24.18 & 21.13  \\
Factorized (rare first) & UC &  7.35 & 22.19 & 19.22 \\

FCL (rare first) & UC &  {\bf 13.16} & 24.23  & {\bf 22.01} \\
\hline
VCL \cite{hou2020visual} (non-rare first) & UC & 16.22 & 18.52 & 18.06 \\
Baseline (non-rare first) & UC &  13.47 & 19.22 & 18.07 \\
Factorized (non-rare first) &UC& 15.72 & 16.95 & 16.71 \\

FCL (non-rare first) & UC & {\bf 18.66} & {\bf 19.55}  & {\bf 19.37} \\

\hline\hline
FG \cite{bansal2020detecting} &  UO & 11.22 &  14.36 & 13.84  \\
Baseline & UO & 12.86 & 20.77 & 19.45  \\
FCL &  UO & {\bf 15.54} &  20.74  & {\bf 19.87} \\
\hline
\end{tabular}
\end{table}

\begin{table}[tp]
\caption{Comparison to the state-of-the-art approaches on HICO-DET dataset~\cite{chao2018learning}. FCL $^{DRG}$ is FCL with object detector provided by \cite{gao2020drg}. FCL + VCL means we fuse the result provided in \cite{hou2020visual} with FCL. VCL$^{DRG}$ uses the released model of VCL.}
\label{table:sota_hico}
\centering
\resizebox{0.95\linewidth}{!}{
\begin{tabular}{@{}lcccccc@{}}
\hline
\multirow{2}{*}{Method} &
\multicolumn{3}{c}{Default}&\multicolumn{3}{c}{Known Object}\cr\cline{2-7}
&Full&Rare&NonRare&Full&Rare&NonRare\cr
\hline\hline
FG \cite{bansal2020detecting}  & 21.96 & 16.43 & 23.62 & - & - & - \\
IP-Net \cite{wang2020learning} & 19.56 & 12.79 & 21.58 & 22.05 & 15.77 & 23.92 \\
PPDM \cite{liao2019ppdm} &  21.73 & 13.78 &24.10 &24.58 &16.65 &26.84 \\
VCL \cite{hou2020visual} & 23.63 & 17.21 & 25.55 & 25.98 & 19.12 & 28.03 \\
DRG \cite{gao2020drg} & 24.53 & 19.47 & 26.04 & 27.98 & 23.11 & 29.43 \\

\hline

Baseline & 23.35 & 17.08 & 25.22 & 25.44 & 18.78 & 27.43 \\
FCL & {\bf 24.68} & {\bf 20.03 } & {\bf 26.07} & {\bf 26.80} & {\bf 21.61} & {\bf 28.35 }\\
FCL + VCL & {\bf 25.27} & {\bf 20.57} & {\bf 26.67} & {\bf 27.71} & {\bf 22.34} & {\bf 28.93} \\
\hline
VCL \cite{hou2020visual} $^{DRG}$ & 28.33  &  20.69  & 30.62 & 30.59 & 22.40 & 33.04\\
Baseline$^{DRG}$ & 28.12 & 21.07 & 30.23 & 30.13 &22.30 &32.47\\
FCL $^{DRG}$ & {\bf 29.12} & {\bf 23.67} & {\bf 30.75} & {\bf 31.31} & {\bf 25.62} & {\bf 33.02} \\
(FCL + VCL) $^{DRG}$ & {\bf 30.11} & {\bf 24.46} & {\bf 31.80} & {\bf 32.17} & {\bf 26.00} & {\bf 34.02} \\

\hline
VCL \cite{hou2020visual} $^{GT}$ & 43.09 & 32.56 &  46.24 & - & - & - \\
FCL$^{GT}$ & {\bf 44.26} &  {\bf 35.46} & {\bf 46.88}  & - & - & - \\
(FCL + VCL)$^{GT}$ & {\bf 45.25} & {\bf 36.27} & {\bf 47.94} & - & - & - \\
\hline
\end{tabular}
}
\end{table}

{\bf Unseen composition}. Table~\ref{table:zero_shot1} shows that FCL achieves large improvement on Unseen category by {\bf 4.22\%} and {\bf 5.19\%} than baseline, and by {\bf 3.10\% and 2.44\%} compared to previous works~\cite{bansal2020detecting, hou2020visual} on the two selection strategies respectively. Meanwhile, the two selection strategies witness a consistent improvement with FCL on nearly all categories, which indicates that composing novel HOI samples contributes to overcome the scarcity of HOI samples. In rare first selection, FCL has a similr result to baseline and VCL \cite{hou2020visual} on Seen category. But step-wise optimization can improve the result on Seen category and Full category (See Table~\ref{table:step}). In addition, the factorized model has a very poor performance in the head classes compared to our baseline. Noticeably, factorized model achieves better performance on Unseen category than baseline in non-rare first selection while has worse result on Unseen category in rare first selection. FCL witnesses a consistent improvement in different evaluation settings. In the remaining data, unseen HOIs of rare first zero-shot have more rare verbs (less than 10 instances) than that of non-rare first zero-shot.

{\bf Unseen object}. We further evaluate FCL in novel object zero-shot HOI detection, which requires to detect HOIs that is interacting with novel objects. Table~\ref{table:zero_shot1} shows FCL effectively improves the baseline by 2.68\% on Unseen Category, although there are no real objects of unseen HOIs in training set. This illustrates the ability of FCL for detecting unseen HOIs with novel objects. Here, the same as \cite{bansal2020detecting}, we also use a generic detector to enable unseen object detection.





\subsubsection{Effectiveness for Long-Tailed HOI Detection}

We compare FCL with
recent state-of-the-art HOI detection approaches \cite{wang2020learning, liao2019ppdm, bansal2020detecting, hou2020visual, gao2020drg} using fine-tuned object detector on HICO-DET to validate its effectiveness on long-tailed HOI detection. For fair comparison, we use the same fine-tuned object detector provided by \cite{hou2020visual}. For evaluation, we follow the settings in \cite{chao2018learning}: Full (600 HOIs), Rare (138 HOIs), Non-Rare (462 HOIs) in ``Default" and ``Known Object" on HICO-DET.

In Table~\ref{table:sota_hico}, we find that the proposed method achieves new state-of-the-art performance, {\bf 24.68\%} and {\bf 26.80\%} mAP on ``Default" and ``Known Object". Meanwhile, we achieve a significant performance improvement of {\bf 2.82\%} over the contemporary best rare performance model \cite{hou2020visual} under the same object detector, which indicates the effectiveness of the proposed compositional learning for the long-tailed HOI detection. Furthermore, with the same object detection result to \cite{gao2020drg}, our results surprisingly increase to {\bf 29.12\%} on ``Default'' mode. Here, we merely change the detection result provided in \cite{hou2020visual} to that provided in \cite{gao2020drg} during inference.
Particularly, we find our method is complementary to compose HOIs between images \cite{hou2020visual}. By simply fusing the result provided by \cite{hou2020visual} with FCL, we can further largely improve the results under different object detectors.

\begin{table}[tp]
\centering
\small
 \caption{Illustration of proposed modules under step-wise optimization. FCL means proposed Fabricated Compositional Learning. V indicates the verb regularization loss.}
 \begin{tabular}{@{}ccccccc@{}}
 
\hline
FCL & V &Full&Rare&NonRare&Unseen\cr
\hline\hline

\hline
- & - &  18.12 & 15.99 & 20.65 & 12.41 \\
\checkmark & - & 19.08 & 17.47 & 20.95 & 14.90\\
- &\checkmark &  18.32 & 16.73 & 20.82 & 12.23 \\
\checkmark & \checkmark & {\bf 19.61} & {\bf 18.69} & {\bf 21.13} &{\bf 15.86} \\


\hline
\end{tabular}

\label{table:ablation}

\end{table}

\begin{table}
\small
\centering
\caption{Ablation study of fabricator under step-wise optimization. FCL within image means we compose HOIs within image. + verb fabricator is we fabricate verb and object features.}
 \begin{tabular}{@{}cccccc@{}}
\hline
Method &Full&Rare&NonRare&Unseen\cr
\hline\hline

\hline

FCL & {\bf 19.61} & {\bf 18.69} & 21.13 &{\bf 15.86} \\
FCL w/o noise  & 19.45 & 17.69 & 21.22 & 15.74 \\
FCL w/o verb & 19.20 & 18.02 & 21.04 & 14.71 \\
%
FCL + verb fabricator & 19.47 & 16.93 & 21.43 & 15.89 \\


\hline
 \end{tabular}


\label{table:ab_fabricator}

\end{table}

\subsubsection{Effectiveness on V-COCO}
We also evaluate FCL on V-COCO. Although the data on V-COCO is balanced, FCL still improves the baseline (reproduced PMFNet \cite{wan2019pose}) in Table~\ref{table:vcoco}.

\begin{table}
\small
\centering
\caption{Illustration of Fabricated Compositional Learning on V-COCO based on PMFNet \cite{wan2019pose}}
\begin{tabular}{@{}lc@{}}
\hline

Method & $AP_{role}$\cr

\hline\hline

\hline
PMFNet \cite{wan2019pose} & 52.0 \\
Baseline & 51.85\\
FCL & {\bf 52.35} \\
\hline
\end{tabular}

\label{table:vcoco}
\end{table}

\subsection{Ablation Studies}
\label{sec:ab}

For a robust validation of the proposed method in rare categories and unseen categories simultaneously, we select 24 rare categories and 96 non-rare categories for zero-shot learning (remained 30,662 training instances). This result is roughly between non-rare first selection and rare first selection in Table~\ref{table:zero_shot1}. See supplementary material for unseen type details and ablation study of long-tailed HOI detection based on Table~\ref{table:sota_hico}. We conduct ablation study on FCL, verb regularization loss, verb fabricator, step-wise optimization and the effect of object detector.




{\bf Fabricated Compositional Learning}. In Table~\ref{table:ablation}, we find that the proposed compositional method with fabricator can steadily improve the performance and it is orthogonal to verb feature regularization (verb regularization loss).

{\bf Verb Feature Regularization}. We use a simple auxiliary verb loss to regularize verb features. Although verb regularization loss can slightly improve the rare and unseen category performance (See row 1 and row 3 in Table~\ref{table:ablation}), FCL further achieves better performance. This indicates that regularizing factor features is suboptimal compared to the proposed method. Semantic verb regularization like \cite{xu2019learning} has a similar result (See supplementary materials).

{\bf Verb and Noise for Fabricator}. Table~\ref{table:ab_fabricator} demonstrates that performance drops without verb representation or noise. This shows verb representations can provide useful information for generating objects and noise efficiently improves the performance by increasing feature diversity. We meanwhile find the fabricator still effectively improves the baseline without verb or noise by comparing Table~\ref{table:ablation} and Table~\ref{table:ab_fabricator}, which indicates the efficiency of FCL.

{\bf Verb Fabricator}. The result of fabricating verb features (from verb identity embedding, object features and noise) is even worse as in Table~\ref{table:ab_fabricator}. This verifies that it is difficult to directly generate useful verb or HOI samples due to the complexity and abstraction. Supplementary materials provide more visualized analysis of verb and object feature.

\begin{table}[tp]
\caption{Comparison between step-wise optimization and one step optimization. ZS is the setting in our ablation study.}
\label{table:step}
\centering
\small
\begin{tabular}{@{}lcccccc@{}}
\hline
Method &Full&Rare&NonRare&Unseen\cr

\hline\hline

one step (long-tailed) & 24.03 & 18.42 & 25.70 & - \\
step-wise  (long-tailed) & {\bf 24.68} & {\bf 20.03} & {\bf 26.07} & -\\

\hline
one step (ZS) & {\bf 19.69} & 18.22 & 20.82 & {\bf 17.64} \\

step-wise  (ZS) & 19.61 & {\bf 18.69} & {\bf 21.13} & 15.86 \\
\hline
one step (rare first ZS) & 22.01 & 15.55 & 24.56 & {\bf 13.16} \\

step-wise  (rare first ZS) & {\bf 22.45} & {\bf 17.19} & {\bf 25.34} & 12.12 \\

\hline
one step (non-rare ZS) & {\bf 19.37} & 15.39 & 20.56 & {\bf 18.66} \\
step-wise  (non-rare ZS) & 19.11 & {\bf 17.12} & {\bf 21.02} & 15.97 \\
\hline
\end{tabular}
\end{table}

\begin{table}[tp]

\centering
\small
\caption{Illustration of the effect of fine-tuned detectors on FCL. The COCO detector is trained on COCO dataset provided in \cite{wu2019detectron2}.
We fine-tune the ResNet-101 Faster R-CNN detector based on Detectron2 \cite{wu2019detectron2}. Here, the baseline is our model without fabricator. The last column is object detection result on HICO-DET test.
}
\label{table:detector}
\begin{tabular}{@{}cccccc@{}}
\hline
Method & Detector & Full & Rare & NonRare & Object mAP\cr

\hline\hline

\hline
Baseline & COCO & 21.24 & 17.44 & 22.37 & 20.82\\
FCL & COCO & {\bf 21.80} & {\bf 18.73} & {\bf 22.71} & 20.82 \\

\hline
Baseline & HICO-DET & 23.94 & 17.48 & 25.87 & 30.79 \\
FCL & HICO-DET & {\bf 24.68} & {\bf 20.03} & {\bf 26.07} & 30.79\\
\hline
Baseline & GT & 43.63 & 34.23 & 46.43 & 100.00 \\


FCL  & GT & {\bf 44.26} & {\bf 35.46} & {\bf 46.88} & 100.00 \\

\hline

\hline
\end{tabular}
\end{table}

\begin{figure*}
\centering
\includegraphics[width=0.84\textwidth]{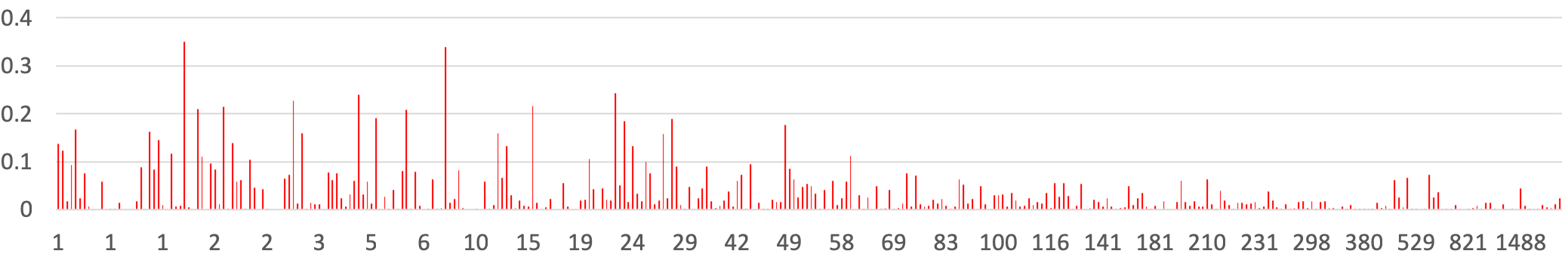}

  \caption{Illustration of the improvement in those improved categories between FCL and baseline on HICO-DET dataset under default setting. The graph is sorted by the frequency of category samples and the horizontal axis is the number of training samples for each category. The result is reported in mAP (\%). The details of category name are provided in supplementary materials.}
\label{fig:long_tailed_improve}
\end{figure*}

{\bf Step-wise Optimization.} Table~\ref{table:step} illustrates that step-wise training has better performance in rare and non-rare categories while has worse performance in unseen categories. We think it might be because the model with the step-wise training has the bias to seen categories in the first step since there are no training data for unseen categories. 


{\bf Object Detector.} The quality of detected objects has important effect on two-stage HOI Detection methods \cite{hou2020visual}. Table~\ref{table:detector} shows that the improvement of FCL over baseline is higher with the fine-tuned detector on HOI data. COCO detector without finetuning on HICO-DET contains a large number of false positive and false negative boxes on HICO-DET due to domain shift, which is in fact less useful to evaluate the effectiveness of modeling human interactions for HOI detection. If the detected boxes during inference are false, the features extracted from the false boxes are also unreal and have large shift to the fabricated objects during training. This causes that fabricated objects are less useful for inferring HOIs during inference. Besides, GT boxes provide a strong object label prior for verb recognition.

\begin{figure}
\centering
\includegraphics[width=0.34\textwidth]{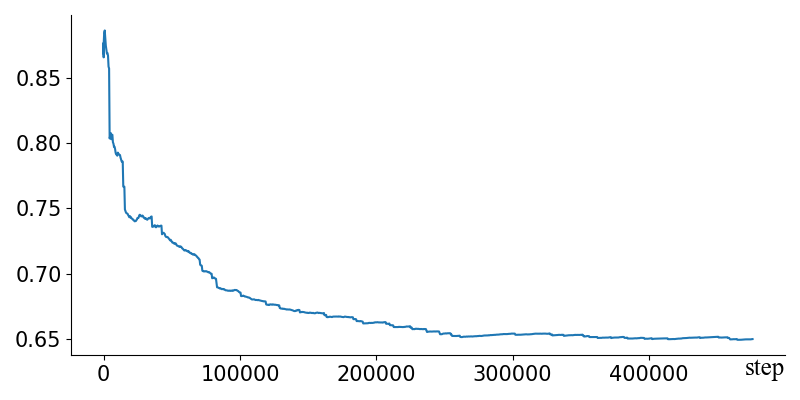}
  \caption{The changing trend of cosine similarity between fabricated object features and real object features during optimization in long-tailed HOI detection in step-wise training.}
\label{fig:cosdis_zero_shot_obj}
\end{figure}

%
%


\section{Qualitative Analysis}
\label{sec:qua}


{\bf Illustration of improvement among categories}. In Figure~\ref{fig:long_tailed_improve}, we find that \textit{the rarer the category is, the more the proposed method can improve}. The result illustrates the benefit of FCL for long-tailed issue in HOI Detection. 


{\bf Visualized Analysis between fabricated and real object features}. Figure~\ref{fig:cosdis_zero_shot_obj} presents that cosine similarity between fabricated and real object features gradually goes down to stability in step-wise training. This demonstrates the end-to-end optimization with shared HOI classifier helps fabricate efficient and similar objects during optimization process. \textit{More analysis of generated object representations by t-SNE is provided in Supplementary Materials}.



\section{Conclusion}
In this paper, we introduce a Fabricated Compostional Learning approach to compose samples for open long-tailed HOI Detection. Specifically, we design an object fabricator to fabricate object features, and then stitch the fake object features and real verb features to compose HOI samples. Meanwhile, we utilize an auxiliary verb regularization loss to regularize the verb feature for improving Human-Object Interaction generalization. Extensive experiments illustrate the efficiency of FCL on the largest HOI detection benchmarks, particularly for low-shot and zero-shot detection.

\noindent {\bf Acknowledgements} This work was supported in part by Australian Research Council Projects FL-170100117, DP-180103424, IH-180100002, and IC-190100031.

{\small
\bibliographystyle{ieee_fullname}
\bibliography{egbib}
}

\begin{appendices}

\newpage

\section{Overview of Appendixes}
\label{sec:ab}
In this supplementary file, we provide additional details of the proposed method in Section B. Section C demonstrates more quantitative analysis (\eg Object identity embedding, VRD, Semantic verb regularization, comparison of detector, additional ablation study and so on). In the last section, we illustrate the qualitative results (\eg analysis of fabricated features).


\section{Additional Details of the Proposed Method}
\label{sec:method}

\subsection{More examples of Open Long-tailed HOI Detection}
\begin{figure*}[t]
\begin{center}
\includegraphics[width=0.95\textwidth]{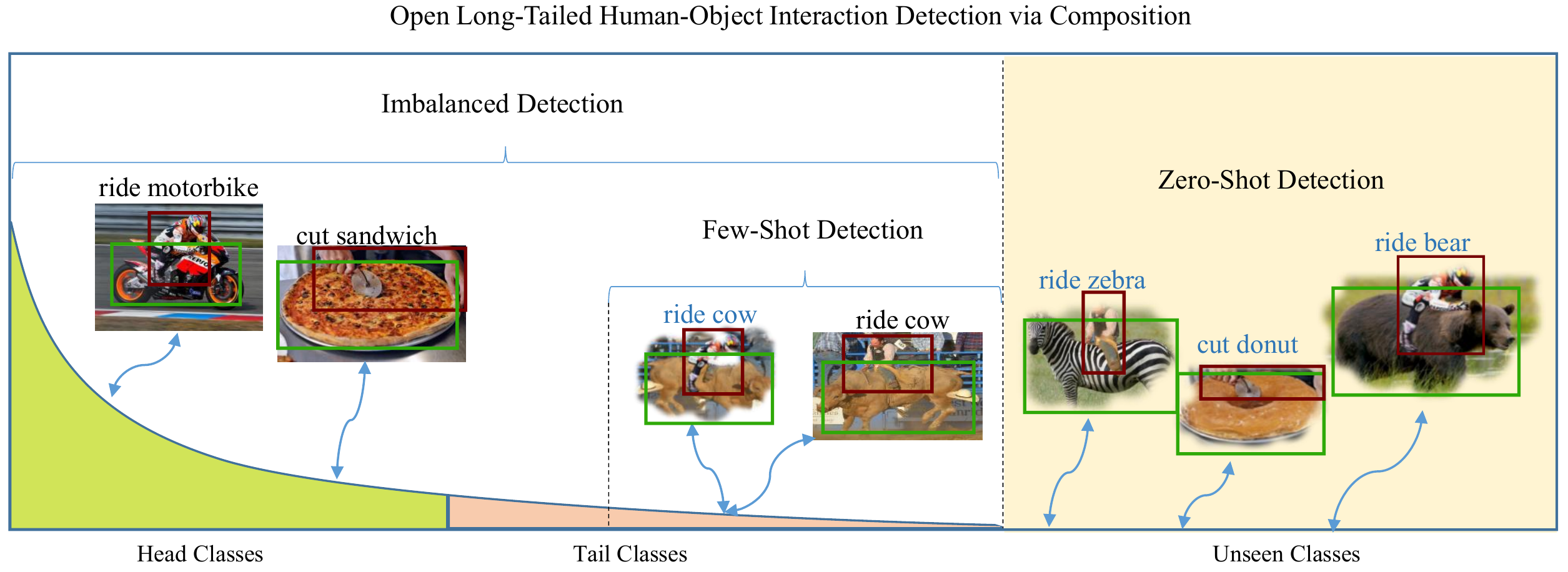}
\end{center}
   \caption{Open long-tailed HOI detection addresses the problem of imbalanced learning and zero-shot learning in a unified way. We propose to compose new HOIs for open long-tailed HOI detection. Specifically, the blurred HOIs, \eg, ``ride bear", are composite, while the black HOIs are real.}
\label{fig:open_long_tailed_large}
\end{figure*}

Figure~\ref{fig:open_long_tailed_large} provides more clear illustration of open long-tailed HOI detection. Open long-tailed HOI detection aims to detect head, tail and unseen classes in one integrated way from long-tailed HOI examples. 

\begin{figure*}
\centering
\includegraphics[width=.95\textwidth]{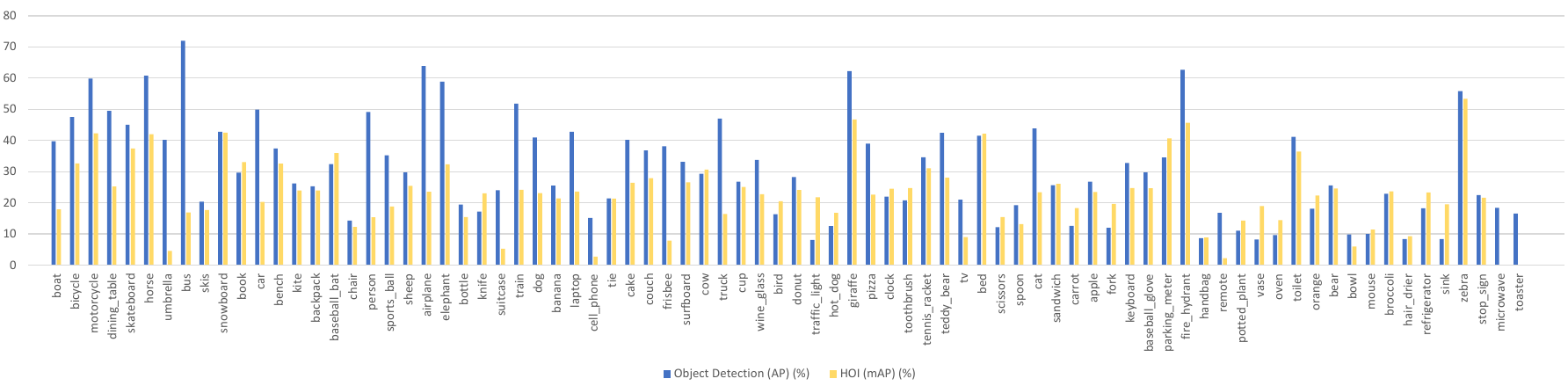}
  \caption{Illustration of Object detection result and HOI detection result in HICO-DET dataset. Blue is Object result. Yellow is HOI result. We average HOI detection AP according to the object categories for a direct comparison.}
\label{fig:obj_hoi_map}
\end{figure*}

\subsection{Factorized model}

We implement the factorized model under our framework. In details, we replace the HOI branch in Figure 3 in the paper with verb and object stream. The two streams predict the verb and object respectively. During inference, we merge the score of verb and object to obtain HOI score as follows,

\begin{equation}
\label{eq:score}
\mathbf{S}_{hoi} = (\mathbf{S}_o \mathbf{A}_o) + (\mathbf{S}_v\mathbf{A}_v),
\end{equation}

where $\mathbf{A}_v$ ($\mathbf{A}_o$) is the co-occurrence matrix between verbs (objects) and HOIs, $\mathbf{S}_o$ is the score from object stream and $\mathbf{S}_v$ is the score from verb stream.

\subsection{The Effect of Objects on HOI Detection}
In the nature, different types of objects form a long-tail distribution. Then, all those actions that people perform on those objects are inevitably long-tailed. As a result, those HOIs that we observed are long-tailed. This motivates us to fabricate balanced objects for composing HOI samples with visual verbs. We have demonstrated the long-tailed distribution of objects in Figure 2 in the paper and the effect of different object detector on HOI detection in Table 7 in paper. We further illustrate HOI detection has roughly similar performance to object detection among most object categories in Figure~\ref{fig:obj_hoi_map}, which also illustrates the importance of object detector for HOI detection at the same time. Meanwhile, it is necessary to balance the the distribution of objects.

\subsection{The Number of Primitives in two Zero-Shot Setting}
We have count the number of unseen HOI primitives (\ie verb and object) in the remaining data of two zero-shot setting. Unseen HOIs of rare first zero-shot has 40 verbs, 5 of which have less than 10 instances in the remaining data, while Unseen HOIs of non-rare first zero-shot have only 30 verbs and all have more 10 instances. We think this partly explains why Factorized method has worse result on unseen category in rare first setting. When the primitives of unseen HOI are few in the training data. Factorized method possibly achieves worse result on unseen category.

\subsection{Fusion of HOI prediction and Generic Object Detector}
In our experiment, we directly predict 600 HOI classes in HICO-DET. The predictions of HOI (verb-object pair) also contain object information. We think the object information in HOI prediction and the generic object detector might be complementary. Thus, we convert HOI scores $S_{hoi}$ to object scores and fuse it with $s_o$ as follow,

\begin{equation}
\label{eq:convert_obj}
\hat{s}_o= \beta_1 \frac {(S_{sp} \cdot S_{hoi}) \mathbf{A}_o ^\mathsf{T}}{\mathbf{B}} + \beta_2 s_o,
\end{equation}

Where $\beta_1$ and $\beta_2$ are 0.3 and 0.7 respectively, $\mathbf{B}\in R^{N_o}$ and $\mathbf{B}_i=\sum_{j=0}^{C} \mathbf{A}_{o_{i,j}}$. Then, we use the new object score $\hat{s}_o$ in Equation~\ref{eq:final_score}. Meanwhile, we can also update the object category according to $\hat{s}_o$. Table~\ref{table:obj_score_fusion} shows we can improve the result a bit under VCL detector which provides all scores for each object category. Noticeably, our baseline under VCL detector also uses this strategy and we do not use this in zero-shot settings. For the DRG object detector, we also do not use this strategy. To some extent, this slightly shows \textit{ HOI prediction and object detection can be mutually promoted}, and provides some insights for our future work although this strategy is not much useful.

\begin{table}[tp]
\caption{Illustration of the effect of fusing HOI predicition to object score. This experiment is based on word-embedding object identity FCL model. }
\label{table:obj_score_fusion}
\begin{center}

\begin{tabular}{@{}lcccccc@{}}
\hline
Method &Full&Rare&NonRare\cr
\hline\hline

\hline
FCL$^{VCL}$ w/o Fusion & 24.42 & 19.68 & 25.84\\
FCL$^{VCL}$ & 24.68 & 20.03 & 26.07 \\

\hline
\end{tabular}
\end{center}
\end{table}

\section{Additional Quantitative analysis}

\subsection{Object Identity}
In Table~\ref{table:obj_id}, we compare three kinds of object identity. The object variables are identified after we fine-tune the fabricator in the first step. Meanwhile, in the end-to-end optimization, the object variables can maintain object semantic information. We find word embedding \cite{pennington2014glove} and object variables achieve similar performance ( 24.78\% vs 24.68\%), while the performance of one-hot representation is a bit worse. Particularly, the HOI model is initialized with a pretrained object detector model. Thus, one-step optimization can also optimize the Fabricator according to the pre-trained backbone. In the main paper, the result of long-tailed HOI detection is the model using word embedding as identity embedding. For simplicity, we use randomly initialized variables as object identity embedding for other model, \ie randomly initialize identity embedding.

\subsection{Visual Relation Detection}

We also present the efficiency of FCL in Predicate Detection on Visual Relation Detection \cite{lu2016visual} in Table~\ref{table:vrd}. Here, we combine subject, predicate and fabricated object to generate novel relation samples \cite{zhan2019exploring}. Table~\ref{table:vrd} illustrates an important improvement on zero-shot predicate detection compared to the state-of-the-art approach with FCL.

\begin{table}[tp]
\caption{Illustration of the effect of different object identity in the proposed fabricator on HICO-DET dataset\cite{chao2018learning}. }
\label{table:obj_id}
\begin{center}

\begin{tabular}{@{}lcccccc@{}}
\hline
Method &Full&Rare&NonRare\cr
\hline\hline

\hline
object variables & {\bf 24.78} & {\bf 20.05} & {\bf 26.19}\\
word embedding & 24.68 & 20.03 & 26.07 \\
one-hot  & 24.38 & 19.49 & 25.84 \\

\hline
\end{tabular}
\end{center}
\end{table}

\begin{table}
\centering
 \begin{tabular}{@{}lcc@{}}
\hline
Method &Zero-Shot & All\cr

\hline\hline
MFURLN \cite{zhan2019exploring}& - & 58.2 \\
MFURLN \cite{zhan2019exploring}* & 25.26 & 57.87 \\
Ours & {\bf 27.31} & {\bf 58.31}\\
\hline
\end{tabular}
\caption{Illustration of Predicate Detection in Visual Relation Detection. Zero-shot means the relation (subject, predicate, object) do not exist in the training data.}
\label{table:vrd}
\end{table}

\subsection{Semantic Verb Regularization}
We also experiment with semantic verb regularization similar to \cite{xu2019learning} with Graph Convolutional Network and verb word embeddings graph. In details, we use the cosine distance loss to regularize the visual verb representation to be similar to the corresponding word embedding. Here, similar to \cite{xu2019learning}, we equally treat same category of verbs among different HOIs as same. Table~\ref{table:semantic} illustrates FCL is orthogonal to semantic regularization. Meanwhile, auxiliary verb loss achieve similar performance compared to semantic verb regularization \cite{xu2019learning}. When we incorporate both semantic regularization and auxiliary verb loss, the improvement is limited. This means verb regularization loss in the paper and semantic verb regularization have similar effect on the model.

\begin{table}[]
\centering
 \begin{tabular}{@{}ccccccc@{}}
\hline
FCL & S & V &Full&Rare&NonRare&Unseen\cr
\hline\hline

\hline

 - &\checkmark & -&  18.22 & 15.69 & 20.74 & 12.98 \\
\checkmark & \checkmark& -  & 19.39 & 17.99 &  21.21 & 14.83 \\
\checkmark & - & \checkmark &  19.61 & 18.69 & 21.13 & 15.86 \\
\checkmark & \checkmark & \checkmark &  19.62 & 18.38 & 21.61 & 14.73 \\

\hline
\end{tabular}
\caption{Illustration of semantic regularization modules based on the ablated setting in paper. FCL Means proposed Compostional Learning. S means semantic regularize loss. V means auxiliary verb loss (verb regularization loss in paper).}
\label{table:semantic}
\end{table}

\subsection{Object Feature Regularization}

{\bf visual object feature regularization.} Object features are usually more discriminative. Meanwhile, we initialize our backbone with the faster-rcnn pre-trained in COCO dataset, which largely helps us to obtain discriminative object features. Thus, it is unnecessary to use auxiliary object loss to regularize object features (See Table~\ref{table:obj}). Meanwhile, we find the object features is more discriminative from the t-SNE graph in Figure~\ref{fig:feats_tsne}.


\begin{table}[tp]
\caption{Illustration of auxiliary object loss on HICO-DET dataset\cite{chao2018learning} based object variables identity. Here, auxiliary object loss aims to regularize visual objects}
\label{table:obj}
\begin{center}

\begin{tabular}{@{}lcccccc@{}}
\hline
Method &Full&Rare&NonRare\cr
\hline\hline

\hline
w/o object loss & {\bf 24.78} & {\bf 20.05} & {\bf 26.19}\\
auxiliary object loss & 24.54 & 19.93 & 25.92 \\
\hline
\end{tabular}
\end{center}
\end{table}


%
\subsection{The Effect of Union Box on FCL}
We extract verb representation from the union box of human and object. In Table~\ref{table:verb_box}, we illustrate with human box verb, FCL still effectively improves the baseline. This shows the proposed method is orthogonal to the verb representation. Noticeably, although the union box contains the object, the HOI model mainly learns the verb representation via compositional learning, and largely ignores the identity information of the object. Thus, the object in the union box do not have much effect on Fabricator. By comparing human box and union box for verb representation in Table 2 in paper and Table~\ref{table:verb_box}, we find verb representation from union box largely improves the performance since it provides more context information for verb representation.

\begin{table}[tp]
\caption{Illustration of the box for verb representation on HICO-DET dataset\cite{chao2018learning}.}
\label{table:verb_box}
\begin{center}

\begin{tabular}{@{}lcccccc@{}}
\hline
Method &Full&Rare&NonRare\cr
\hline\hline

baseline(human box) & 22.91 & 16.66 & 24.77\\
FCL (human box) & 23.83 & 18.62 & 25.39\\
%
\hline
\end{tabular}
\end{center}
\end{table}

\begin{table}[tp]
\caption{The result while filtering out the composite HOIs according to the similarity between the fake objects and original objects. \#Neighbors ($K$) means top $K$ neighbors according to similarity. This experiment is based on ablated setting in Table 3 in paper. When the number of neighbors is 80, we do not filter out composite HOIs according to similarity.}
\label{table:emb_filter}
\begin{center}
\small
\begin{tabular}{@{}lcccccc@{}}

\hline
\#Neighbors ($K$) & 1 & 5 & 10 & 20 & 40 & 80\cr
\hline\hline
FCL (Full) & 18.70 & 19.15 & 19.19 & 19.48 & 19.60 & 19.61 \\
%
\hline
\end{tabular}
\end{center}
\end{table}

\begin{table}[tp]
\caption{Comparison between step-wise optimization and one step optimization in unseen object HOI detection.}
\label{table:step1}
\begin{center}

\begin{tabular}{@{}lcccccc@{}}
\hline
Method &Full&Rare&NonRare&Unseen\cr

\hline\hline

\hline
one step & 19.87 &  15.01 & 22.51 & {\bf 15.54}  \\

step-wise  & {\bf20.13} & {\bf16.71} &{\bf22.82} &13.85\\
\hline
\end{tabular}
\end{center}
\end{table}
\subsection{Additional Object Detector Analysis}
We notice there is a large gap between VCL \cite{hou2020visual} detector and DRG \cite{gao2020drg}. VCL provides the detection result (\ie 30.79\% mAP), while we do not know the detection result of DRG detector. We do not achieve the similar object detection performance to DRG \cite{gao2020drg} when we fine-tune Faster R-CNN on HICO-DET training set. However, we think we can compare the two detector by the recall of HOI detection as illustrated in Table~\ref{table:detector_recall}. Recall can also be used to compare the object detection performance between one-stage HOI detection and two-stage HOI detection. Table~\ref{table:detector_recall} shows FCL$^{DRG}$ nearly achieves similar result to FCL$^{GT}$ on Recall. FCL$^{GT}$ still requires the network to discriminate which pair of human and object boxes has interaction.

\setlength{\tabcolsep}{4pt}
\begin{table}

\centering
\caption{Illustration of recall of HOI under DRG detector, VCL detector and GT boxes.}
\label{table:detector_recall}
\begin{tabular}{@{}lcc@{}}
\hline

Detector &Full (mAP) & Recall (mRec) \cr
\hline\hline
\hline
FCL$^{VCL}$ & 24.68 & 62.07\\ 
FCL$^{DRG}$ & 29.12 & 82.81\\
FCL$^{GT}$ & 44.26 & 86.08 \\  
\hline
\end{tabular}

\end{table}

\subsection{Verb Analysis}
The same verb might has different meanings in different HOIs. However, the verb in HOI dataset (e.g. HICO-DET) mainly represents action. Thus, the verb in HOI dataset is usually not ambiguous. Meanwhile, the deep convolutional network (\eg Resnet) is able to fit some ambiguous and even random data \cite{2017Understanding}. Therefore, we can use factorized method \cite{xu2019learning} for HOI detection and the ambiguous verbs do not affect the compositional learning on HICO-DET \cite{hou2020visual}, even if there are still some ambiguous verbs (e.g. hold) who can be related to multiple objects.

\begin{figure*}

\centering
\includegraphics[width=.95\textwidth]{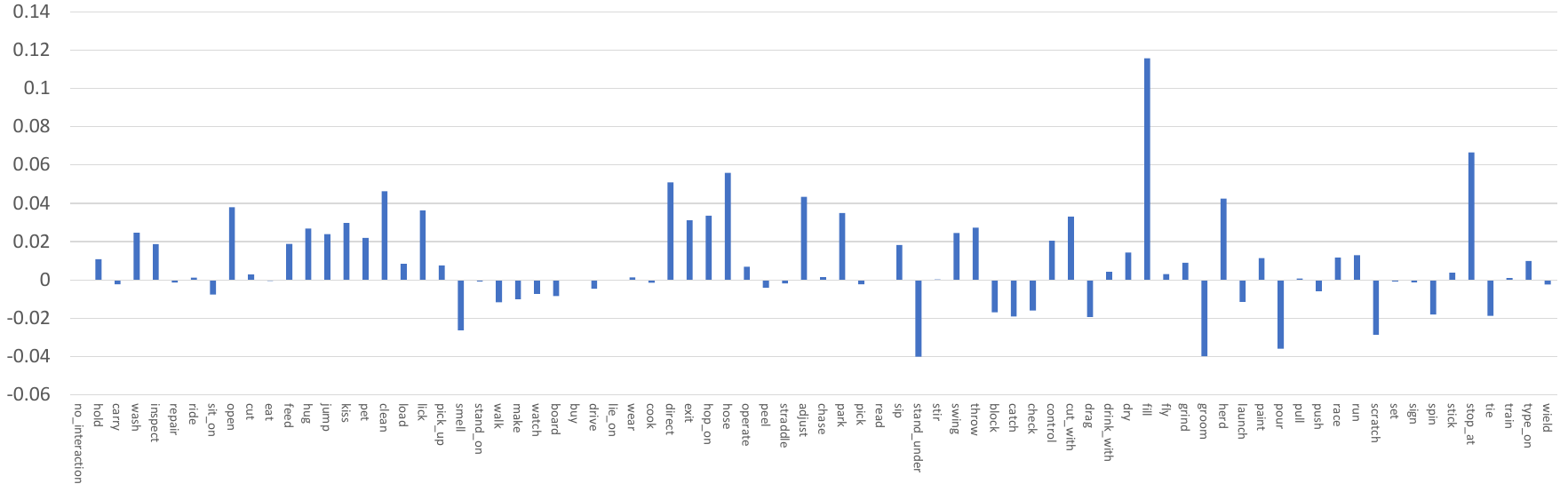}
  \caption{The improvement among the classes of verbs on HICO-DET. The verbs are sorted by the number of HOIs that the particular verb is related. The clear figure is in the directory of Compressed package.}
\label{fig:verb_impro}
\end{figure*}

Besides, we further demonstrates the improvement of FCL among different categories of verbs in Figure~\ref{fig:verb_impro}. We find the ambiguity does not affect the performance of those verbs in fact. For example, although the verb ``hold'' is related to 61 kinds of objects in HICO-DET, the correpsonding HOIs of ``hold'' still achieves considerable improvement.


Inspired by that people interact similar objects in a similar manner. we also design an approach to select composite HOIs according to the similarity between different object of objects, \ie we only keep those composite HOIs whose object is in the top $K$ neighbors of the verb's original object. The original object of the verb is the visual object paired with the verb in the HOI annotation. This helps us to filter out those ambiguous composite HOIs. Specifically, we calculate the similarity between different classes of objects by its word embedding \cite{pennington2014glove}. Then we can obtain the top $K$ neighbors for each class of objects. Table~\ref{table:emb_filter} shows with more similar objects, the performance steadily improves. Particularly, there are only one verb relating to more than 40 HOIs, and 4 verbs with more than 20 HOIs in HICO-DET. When $K=1$, we only keep composite HOIs whose objects have the same label to the original object.

\subsection{Orthogonality to previous methods}

{\bf Orthogonal to spatial pattern}. Table~\ref{table:sp} illustrates that the spatial pattern strategy \cite{gao2018ican, li2018transferable, wan2019pose} largely improves the performance, and the proposed compositional learning is orthogonal to spatial pattern.

{\bf Orthogonal to re-weighting}. In our baseline, we utilize the re-weighting strategy that is used in \cite{li2018transferable, hou2020visual} to compare directly with \cite{hou2020visual}. We demonstrate FCL is orthogonal to re-weighting in Table~\ref{table:wo_rewighting}. Without the useful re-weighting strategy, FCL still achieves similar improvement than baseline.

\setlength{\tabcolsep}{4pt}
\begin{table}

\centering
\caption{Illustration of FCL without re-weighting on long-tailed HOI detection. }
\label{table:wo_rewighting}
\begin{tabular}{@{}cccccc@{}}
\hline

FCL &Full&Rare&NonRare\cr
\hline\hline

\hline

 - & 20.79  &   13.19  &   23.06 \\
\checkmark & 21.20 & 15.48 & 22.90 \\

\hline
\end{tabular}

\end{table}

\subsection{Complementary Analysis of fabricator}
In this section, we conduct analysis of fabricator on HOI detection without unseen data (the full long-tailed HOI detection). We witness the similar trend compared to the ablation study in the paper.

\setlength{\tabcolsep}{4pt}
\begin{table}
\centering
\label{table:ablation_appendix}
\caption{Illustration of proposed modules on long-tailed HOI detection. FCL Means proposed Fabricated Compostional Learning. V means verb regularization loss.}
\begin{tabular}{@{}ccccc@{}}

\hline

FCL& V &Full&Rare&NonRare\cr
\hline\hline

- & -  & 23.35 & 17.08 & 25.22 \\
\checkmark & - & 23.86 & 18.16 & 25.56 \\
- &\checkmark  & 23.94 & 17.48 & 25.87 \\
\checkmark & \checkmark & {\bf 24.78} & {\bf 20.05} & {\bf 26.19} \\


\hline
\end{tabular}

\end{table}

\setlength{\tabcolsep}{4pt}
\begin{table}
\centering
\label{table:ab_hallucinator}
\caption{Ablation study of fabricator. Verb fabricator means we fabricate verb features.}
\begin{tabular}{@{}lccccc@{}}

\hline

Method &Full&Rare&NonRare\cr
\hline\hline


FCL & {\bf 24.78} & {\bf 20.05} & {\bf 26.19} \\
FCL w/o noise  & 24.22 & 19.23 & 25.72 \\
FCL w/o verb & 24.29 & 18.98 & 25.87 \\
verb fabricator & 23.93 & 17.10 & 25.97 \\
\hline
\end{tabular}

\end{table}

\begin{table}[]
    \centering
 \begin{tabular}{@{}ccccccc@{}}
\hline
FCL & SP & ZS &Full&Rare&NonRare& Unseen\cr
\hline\hline

- & - & - & 21.07 & 14.11 & 23.15 & - \\
\checkmark & - & -& 21.68 & 16.92 & 23.11  & - \\
\checkmark & \checkmark & - & {\bf 24.78} & {\bf 20.05} & {\bf 26.19}  & - \\
\hline
- & -  & \checkmark& 15.29 & 14.45 & 17.85  & 8.27 \\
\checkmark & - & \checkmark& 16.82 & 16.57 & 18.17 & 12.94  \\
\checkmark & \checkmark & \checkmark& {\bf 19.61} & {\bf 18.69} & {\bf 21.13} &{\bf 15.86} \\

\hline
\end{tabular}
\caption{Illustration of spatial pattern. SP means we use spatial pattern. ZS means zero-shot setting.}
\label{table:sp}

\end{table}

{\bf Verb and Noise for fabricating objects}. Table~\ref{table:ab_hallucinator} demonstrates the efficiency of verb and noise. Particularly, the performance in the full HOI detection drops larger than that in zero-shot study in the paper. We think it is because the improvement on unseen category is large, while there are no unseen category in the full HOI detection.

{\bf Verb Fabricator}. Table~\ref{table:ab_hallucinator} illustrates if we fabricate verb features to augment HOI samples, the performance apparently decreases to 23.93\% in long-tailed HOI detection. This again illustrates that the verb feature is more complex and it is difficult to generate efficient verb features to facilitate HOI detection.

\begin{figure*}

\centering
\includegraphics[width=.95\textwidth]{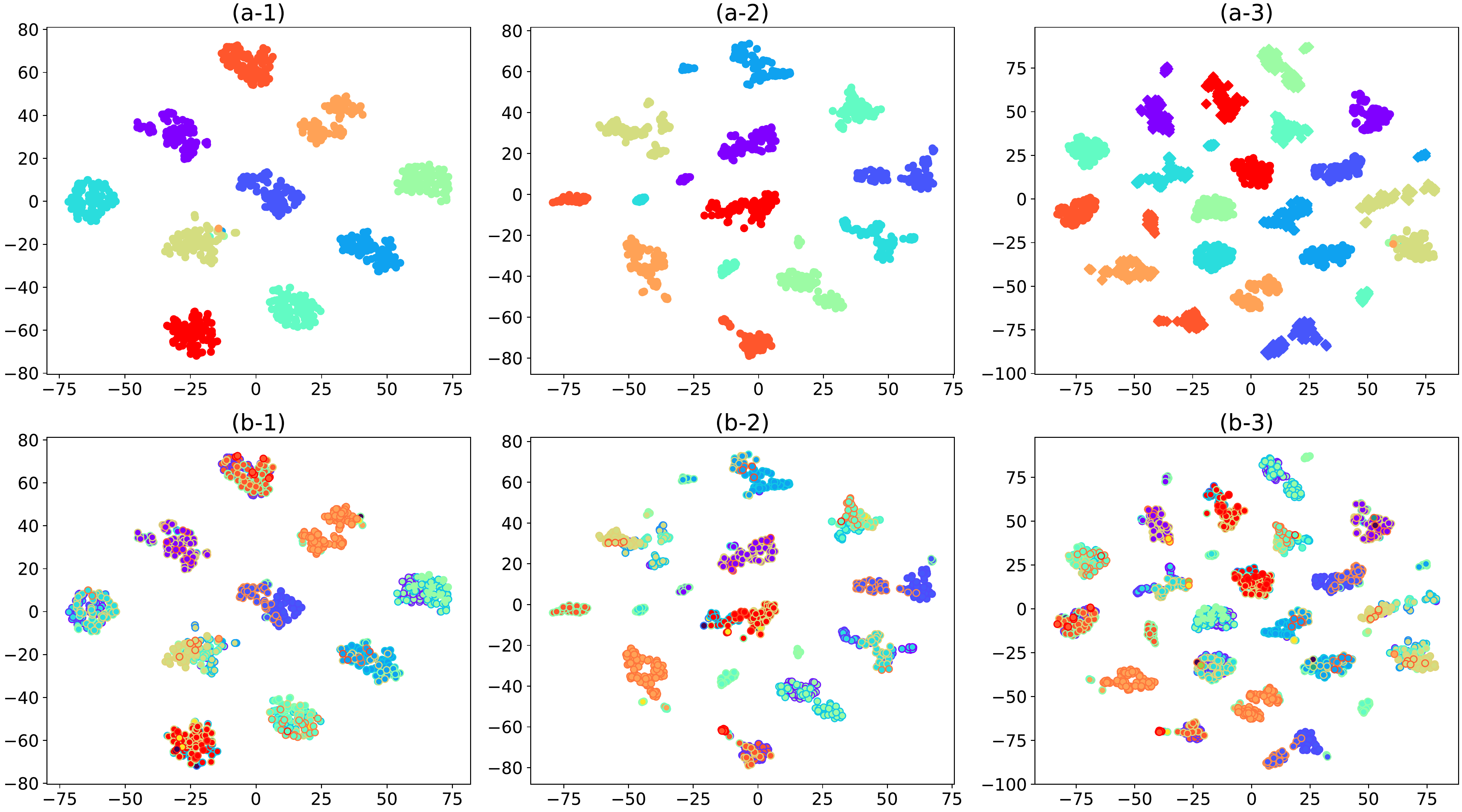}
  \caption{The illustration of real object representations, fabricated object representations and joint representations extracted from long-tailed HOI detection model. We select top 10 frequent object classes from HICO-DET training data. For each classes, we randomly select 100 instances. Column 1 is real object representations, Column 2 is fabricated object representations and Column 3 is the joint representations. In Column 3, diamond point means fabricated object representations. Raw a is the base t-SNE figure. In raw b, we label different verbs with different edges (color) in Raw b.}
\label{fig:feats_tsne_long_tailed}
\end{figure*}



\begin{figure*}

\centering
\includegraphics[width=.95\textwidth]{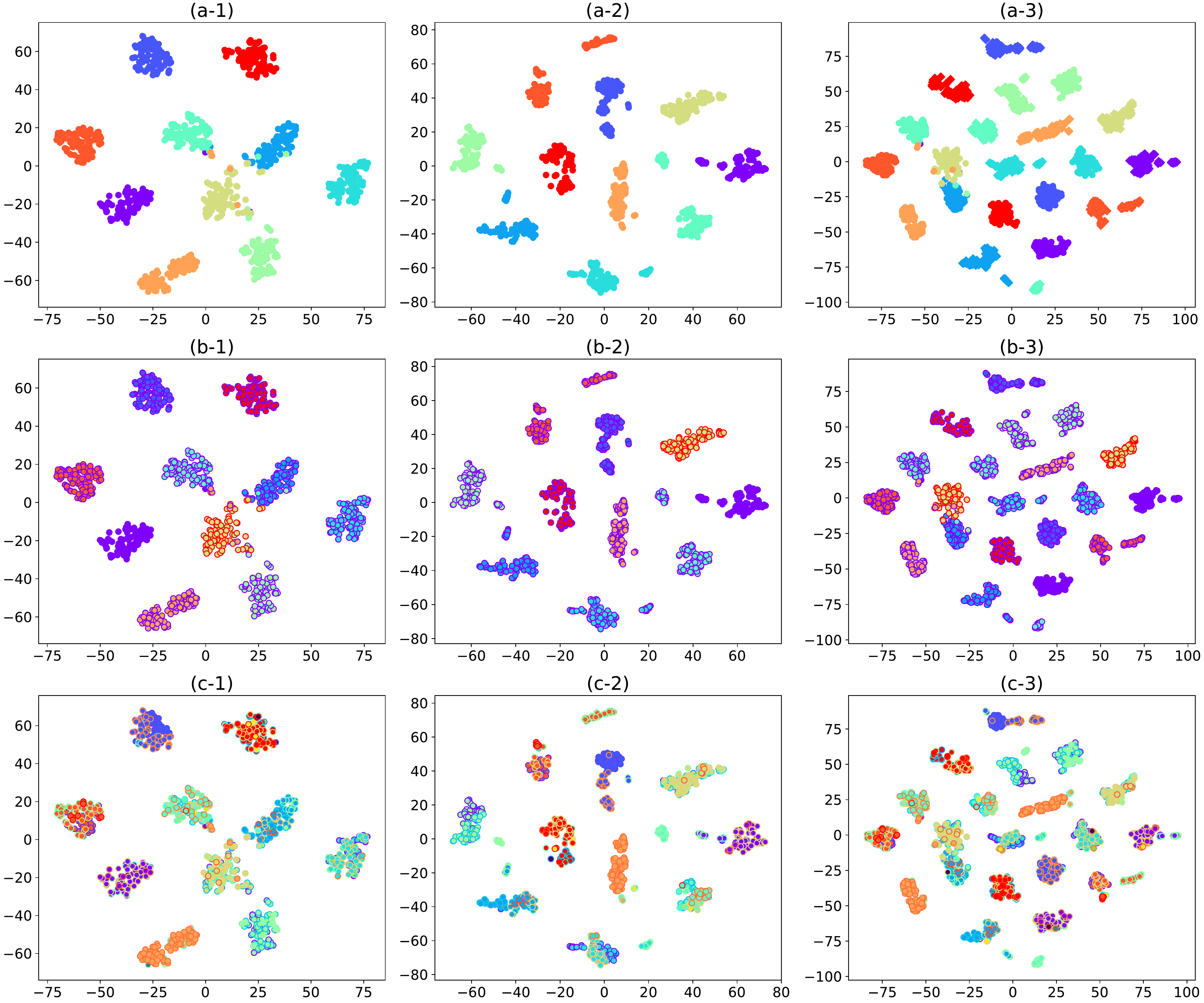}
  \caption{The illustration of real object representations, fabricated object representations and joint representations extracted from unseen object zero-shot model. Column 1 is real object representations, Column 2 is fabricated object representations and Column 3 is the joint representations. In Column 3, diamond point means fabricated object representations. Raw a is the base t-SNE figure. In raw b, we point out the unseen objects with red edge. In Raw c, we label different verbs with different edges (color).}
\label{fig:feats_tsne_zs}
\end{figure*}





\begin{figure*}

\centering
\includegraphics[width=.85\textwidth]{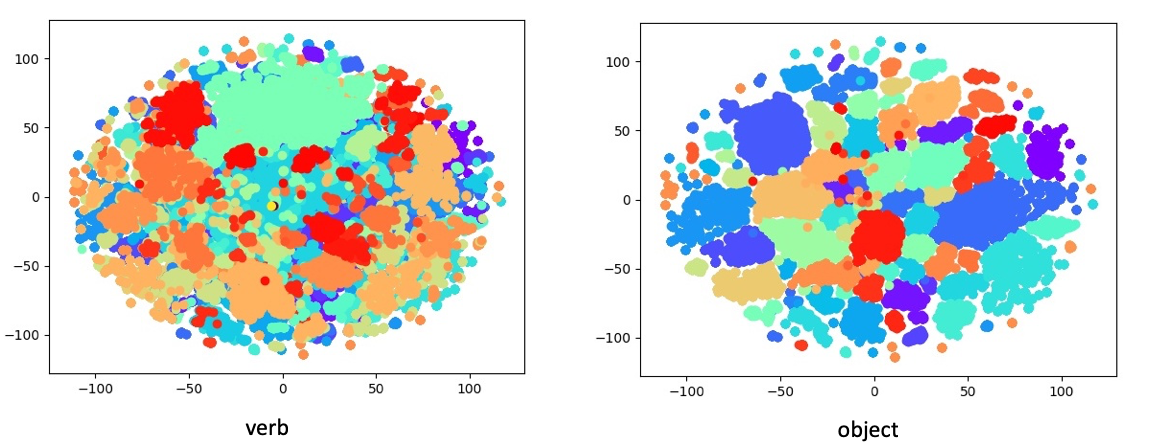}
  \caption{The comparison between verb features and object features.}
\label{fig:feats_tsne}
\end{figure*}

\begin{figure*}

\centering
\includegraphics[width=.85\textwidth]{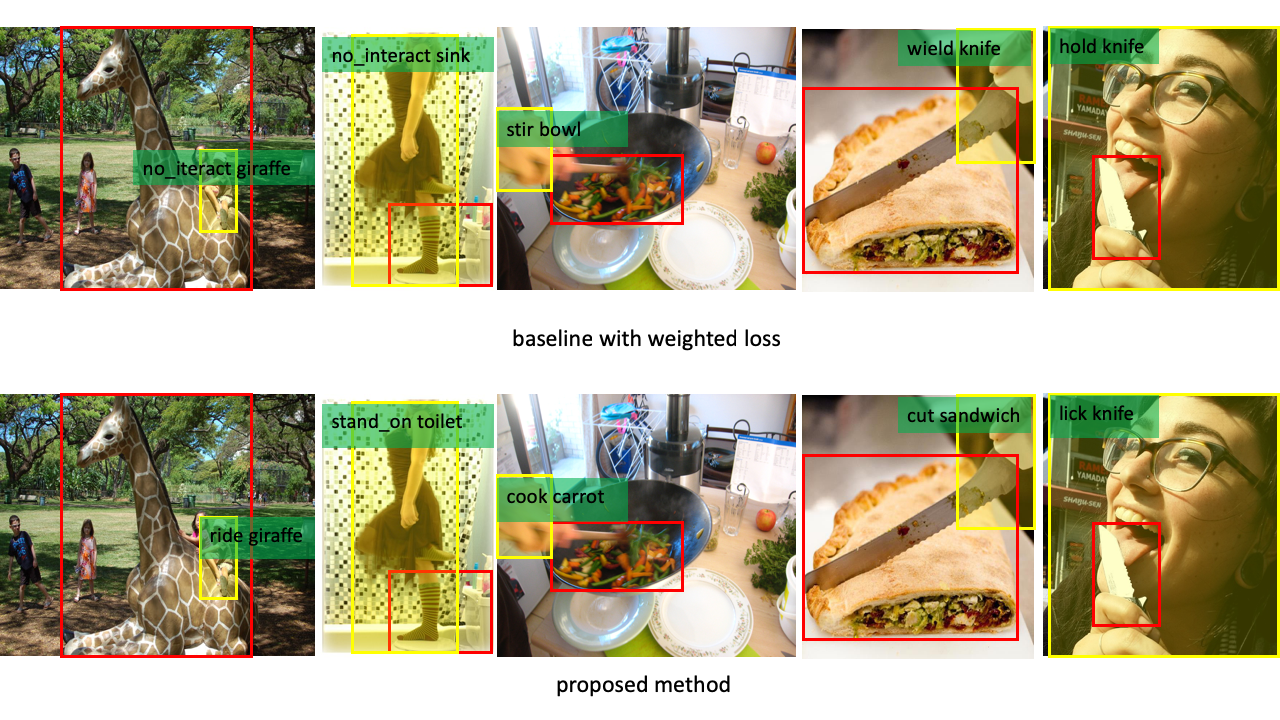}
  \caption{Visual Comparison between FCL and our baseline. The two models use same detector.}
\label{fig:vis_compa}
\end{figure*}

\begin{figure}
\centering
\includegraphics[width=.45\textwidth]{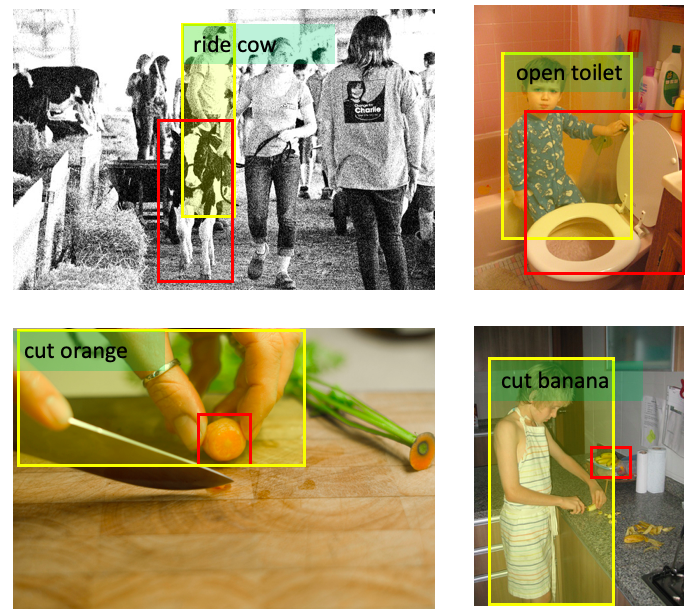}
  \caption{Illustration of failure cases.}
\label{fig:failure_case}
\end{figure}

\subsection{Additional Ablation Study}

{\bf Step-wise optimization}. We also provide the comparison between step-wise optimization and one-step optimization in unseen object HOI detection in Table~\ref{table:step1}.

{\bf Hyper-Parameters}. We follow the hyper-parameters in \cite{hou2020visual} for $\lambda_1$ and $\lambda_2$. For $\lambda_3$, we provide the ablated experiment in Table~\ref{table:ab_hyper} based on 0.5 because we think $L_{reg}$ is less important than $L_{CL}$.

{\bf Fine-tune the network}. In the step-wise optimization, we fine-tune the whole FCL network in the last step. For a fair comparison, we also fine-tune our baseline after we train our network. Table~\ref{table:ab_fine_tune} shows fine-tuning the network improves effectively the baseline. This is the reason why our baseline is strong. It might be because the initial learning 0.01 in our optimization is high.

\setlength{\tabcolsep}{4pt}
\begin{table}[tp]
\caption{Illustration of ablated study on $\lambda_3$ in HICO-DET based on open long-tailed HOI detection (corresponding to Table 3 in paper). }
\label{table:ab_hyper}
\begin{center}

\begin{tabular}{@{}lccc@{}}
\hline
$\lambda_3$ & 0.1 & 0.3 & 0.5\\
\hline\hline
FCL & 19.30& 19.61& 19.10\\

\hline
\end{tabular}
\end{center}
\end{table}

\setlength{\tabcolsep}{4pt}
\begin{table}[tp]
\caption{Ablation study of fine-tuning the network. }
\label{table:ab_fine_tune}
\begin{center}

\begin{tabular}{lccc}
\hline
Method &Full&Rare&NonRare\cr
\hline\hline
Baseline (w/o fine-tune) & 22.83 & 16.32 & 24.77 \\
Baseline & {\bf 23.35} & {\bf 17.08} & {\bf 25.22}\\
\hline
\end{tabular}
\end{center}
\end{table}


\section{Additional Qualitative Analysis}
\label{sec:qua1}

\subsection{Object Representations}
We analyze the real object features and fabricated object features in detail in Figure~\ref{fig:feats_tsne_long_tailed}, ~\ref{fig:feats_tsne_zs} by selecting top 10 frequent classes in HICO-DET. 1) In Figure~\ref{fig:feats_tsne_long_tailed} (a) and Figure~\ref{fig:feats_tsne_zs} (a), we find the fake object features of the same class are close to each other, while the features from different classes are separable although they might share the same verb. 2) Figure~\ref{fig:feats_tsne_long_tailed} (b) and Figure~\ref{fig:feats_tsne_zs} (c) show features of different verbs slightly cluster together within each object class. {\bf We can find there are outliers in some object classes because those outliers have different verbs}. 3) for unseen object ZSL, Figure~\ref{fig:feats_tsne_zs} shows all fake object features of the same class are also closer to each other. Particularly, the unseen objects (red edge in row b) are also separable from others. 4) The Column 3 in Figure~\ref{fig:feats_tsne_long_tailed} and Figure~\ref{fig:feats_tsne_zs} illustrate fake object features are still separable from its real objects of the same class. However, there are still some fabricated features are closer to it's corresponding real features (\eg the dark blue class in Figure~\ref{fig:feats_tsne_long_tailed} and the jade-green class in Figure~\ref{fig:feats_tsne_zs}). We think Column 3 in the two Figures also shows a future direction for fabricating objects, \ie generate more realistic objects.

\subsection{Primitive Features}
Figure~\ref{fig:feats_tsne} illustrates verb features are apparently more difficult to distinguish. The verb representation is abstract and complicated. By contrast, object representations extracted from modern object detector are more discriminative. By comparing Figure~\ref{fig:feats_tsne} with the Figures in VCL \cite{hou2020visual}, we can find the objects of FCL are more discriminative.

\subsection{Qualitative Comparison}
In Figure ~\ref{fig:vis_compa}, we compare our baseline with our proposed method. Apparently, our proposed method efficiently detects rare categories, while the corresponding baseline can not. In fact, all the HOIs detected by our method in Figure~\ref{fig:vis_compa} have less than five samples in training set which is much less than the rare setting (less than 10 samples).

\subsection{Failure cases analysis}
We provide some false positive results on Rare category in Figure~\ref{fig:failure_case}. All failure cases can be separated into four groups: blurry image, wrong verb, wrong object, wrong match. If the image is blurry or has partial occlusion, it is hard to detection the interaction right. Besides, verb is usually hard to classify. Meanwhile, small objects also cause that the network detect object wrongly (\eg the carrot in Figure~\ref{fig:failure_case}). Lastly, even though the network can recognize action and object correctly, it also possibly mismatches the interaction. For example, in Figure~\ref{fig:failure_case}, the women do not interact with the banana on the corner of the table.

\end{appendices}

\end{document}